\documentclass[10pt,twocolumn,letterpaper]{article}

\usepackage{cvpr}
\usepackage{times}
\usepackage{epsfig}
\usepackage{graphicx}
\usepackage{amsmath}
\usepackage{amssymb}

\usepackage{multirow}
\usepackage{rotating}
\usepackage{tabularx}
\usepackage{tikz}
\usetikzlibrary{arrows}
\usepackage{mathtools}
\usepackage{subcaption}
\usepackage{booktabs}
\usepackage{makecell}

\graphicspath{{images/}}

\newcommand{\quotes}[1]{``#1''}
\newcommand{\singlequotes}[1]{`#1'}
\newcolumntype{C}{>{\centering\arraybackslash}m{1cm}}
\newcolumntype{D}{>{\centering\arraybackslash}m{1.2cm}}
\newcolumntype{E}{>{\centering\arraybackslash}p{2.65cm}}
\newcolumntype{F}{>{\centering\arraybackslash}p{2.3cm}}

\makeatletter
\DeclareRobustCommand\bigop[1]{%
  \mathop{\vphantom{\sum}\mathpalette\bigop@{#1}}\slimits@
}
\newcommand{\bigop@}[2]{%
  \vcenter{%
    \sbox\z@{$#1\sum$}%
    \hbox{\resizebox{\ifx#1\displaystyle.7\fi\dimexpr\ht\z@+\dp\z@}{!}{$\m@th#2$}}%
  }%
}
\makeatother
\newcommand{\concat}{\DOTSB\bigop{\lvert\lvert}}
\usepackage[pagebackref=true,breaklinks=true,letterpaper=true,colorlinks,bookmarks=false]{hyperref}

\hypersetup{
    bookmarks=true,         
    unicode=false,          
    pdftoolbar=true,        
    pdfmenubar=true,        
    pdffitwindow=false,     
    pdftitle={Latent Space Autoregression for Novelty Detection},    
    pdfauthor={Davide Abati},     
    pdfsubject={},   
    pdfcreator={Davide Abati},   
    pdfproducer={},  
    pdfkeywords={}, 
    pdfnewwindow=true,      
    colorlinks=false,       
    linkcolor=red,          
    citecolor=green,        
    filecolor=magenta,      
    urlcolor=cyan           
}

\cvprfinalcopy 
\def\cvprPaperID{858} 

\ifcvprfinal\fi

\newif\ifcr
\crtrue
\ifcr
\DeclareGraphicsExtensions{.pdf,.png,.jpg}
\newcommand{\dvd}[1]{}
\newcommand{\ngl}[1]{}
\newcommand{\smn}[1]{}
\newcommand{\rti}[1]{}
\else
\DeclareGraphicsExtensions{.jpg,.png,.pdf}
\newcommand{\dvd}[1]{\textcolor{red}{\textsc{DVD}: #1}}
\newcommand{\ngl}[1]{\textcolor{blue}{\textsc{NGL}: #1}}
\newcommand{\smn}[1]{\textcolor{cyan}{\textsc{SMN}: #1}}
\newcommand{\rti}[1]{\textcolor{magenta}{\textsc{RTI}: #1}}
\fi

\DeclarePairedDelimiterX{\infdivx}[2]{(}{)}{%
  #1\;\delimsize\|\;#2%
}

\usepackage{xr}
\makeatletter
\newcommand*{\addFileDependency}[1]{
  \typeout{(#1)}
  \@addtofilelist{#1}
  \IfFileExists{#1}{}{\typeout{No file #1.}}
}
\makeatother
 
\newcommand*{\myexternaldocument}[1]{%
    \externaldocument{#1}%
    \addFileDependency{#1.tex}%
    \addFileDependency{#1.aux}%
}
\myexternaldocument{supplementary}

\begin{document}

\title{Latent Space Autoregression for Novelty Detection}

\author{Davide Abati \quad Angelo Porrello \quad Simone Calderara \quad Rita Cucchiara\\\\
University of Modena and Reggio Emilia\\
{\tt \small\{name.surname\}@unimore.it}
}

\maketitle
\begin{abstract}
Novelty detection is commonly referred to as the discrimination of observations that do not conform to a learned model of regularity. Despite its importance in different application settings, designing a novelty detector is utterly complex due to the unpredictable nature of novelties and its inaccessibility during the training procedure, factors which expose the unsupervised nature of the problem. In our proposal, we design a general framework where we equip a deep autoencoder with a parametric density estimator that learns the probability distribution underlying its latent representations through an autoregressive procedure. 
We show that a maximum likelihood objective, optimized in conjunction with the reconstruction of normal samples, effectively acts as a regularizer for the task at hand, by minimizing the differential entropy of the distribution spanned by latent vectors. In addition to providing a very general formulation, extensive experiments of our model on publicly available datasets deliver on-par or superior performances if compared to state-of-the-art methods in one-class and video anomaly detection settings. Differently from prior works, our proposal does not make any assumption about the nature of the novelties, making our work readily applicable to diverse contexts.
\end{abstract}
\setlength\parindent{0pt}
\section{Introduction}
Novelty detection is defined as the identification of samples which exhibit significantly different traits with respect to an underlying model of regularity, built from a collection of normal samples.
The awareness of an autonomous system to recognize unknown events enables applications in several domains, ranging from video surveillance~\cite{calderara2011detecting,hasan2016learning}, to defect detection~\cite{kumar2008computer} to medical imaging~\cite{schlegl2017unsupervised}. Moreover, the surprise inducted by unseen events is emerging as a crucial aspect in reinforcement learning settings, as an enabling factor in curiosity-driven exploration~\cite{pathak2017curiosity}.\\
However, in this setting, the definition and labeling of novel examples are not possible. Accordingly, the literature agrees on approximating the ideal shape of the boundary separating normal and novel samples by modeling the intrinsic characteristics of the former. Therefore, prior works tackle such problem by following principles derived from the unsupervised learning paradigm~\cite{cong2011sparse,sabokrou2018adversarially,hasan2016learning,liu2017future,luo2017revisit}. Due to the lack of a supervision signal, the process of feature extraction and the rule for their normality assessment can only be guided by a proxy objective, assuming the latter will define an appropriate boundary for the application at hand.\\
According to cognitive psychology~\cite{barto2013novelty}, novelty can be expressed either in terms of capabilities to \emph{remember} an event or as a degree of \emph{surprisal}~\cite{tribus1961thermostatics} aroused by its observation. The latter is mathematically modeled in terms of low probability to occur under an expected model, or by lowering a variational free energy~\cite{itti2009bayesian}. 
In this framework, prior models take advantage of either parametric~\cite{zong2018deep} or non-parametric~\cite{hinami2017joint} density estimators. 
Differently, remembering an event implies the adoption of a memory represented either by a dictionary of normal prototypes - as in sparse coding approaches \cite{cong2011sparse} - or by a low dimensional representation of the input space, as in the self-organizing maps~\cite{kohonen2012self} or, more recently, in deep autoencoders. Thus, in novelty detection, the remembering capability for a given sample is evaluated either by measuring reconstruction errors~\cite{hasan2016learning,liu2017future} or by performing discriminative in-distribution tests~\cite{sabokrou2018adversarially}.\\
Our proposal contributes to the field by merging remembering and surprisal aspects into a unique framework: we design a generative unsupervised model (i.e., an autoencoder, represented in Fig.~\ref{fig:cover}) that exploits end-to-end training in order to maximize remembering effectiveness for normal samples whilst minimizing the surprisal of their latent representation. This latter point is enabled by the maximization of the likelihood of latent representations through an autoregressive density estimator, which is performed in conjunction with the reconstruction error minimization. We show that, by optimizing both terms jointly, the model implicitly seeks for minimum entropy representations maintaining its remembering/reconstructive power. While entropy minimization approaches have been adopted in deep neural compression~\cite{balle2016end}, to our knowledge this is the first proposal tailored for novelty detection. In memory terms, our procedure resembles the concept of prototyping the normality using as few templates as possible. Moreover, evaluating the output of the estimator enables the assessment of the surprisal aroused by a given sample.
\section{Related work}
\label{sec:related}
\textbf{Reconstruction-based methods.}
On the one hand, many works lean toward learning a parametric projection and reconstruction of normal data, assuming outliers will yield higher residuals.
Traditional sparse-coding algorithms~\cite{zhao2011online,cong2011sparse,lu2013abnormal} adhere to such framework, and represent normal patterns as a linear combination of a few basis components, under the hypotheses that novel examples would exhibit a non-sparse representation in the learned subspace. 
In recent works, the projection step is typically drawn from deep autoencoders~\cite{hasan2016learning}. 
In~\cite{luo2017revisit} the authors recover sparse coding principles by imposing a sparsity regularization over the learned representations, while a recurrent neural network enforces their smoothness along the time dimension. %
In~\cite{sabokrou2018adversarially}, instead, the authors take advantage of an adversarial framework in which a discriminator network is employed as the actual novelty detector, spotting anomalies by performing a discrete in-distribution test. 
Oppositely, future frame prediction~\cite{liu2017future} maximizes the expectation of the next frame exploiting its knowledge of the past ones; at test time, observed deviations against the predicted content advise for abnormality.
Differently from the above-mentioned works, our proposal relies on modeling the prior distribution of latent representations. This choice is coherent with recent works from the density estimation community~\cite{tomczak2017vae,bauer2018resampled}. However, to the best of our knowledge, our work is the first advocating for the importance of such a design choice for novelty detection.\\\\
\textbf{Probabilistic methods.}
A complementary line of research investigates different strategies to approximate the density function of normal appearance and motion features. 
The primary issue raising in this field concerns how to estimate such densities in a high-dimensional and complex feature space.
In this respect, prior works involve hand-crafted features such as optical flow or trajectory analysis and, on top of that, employ both non-parametric~\cite{adam2008robust} and parametric~\cite{basharat2008learning,mahadevan2010anomaly,li2014anomaly} estimators, as well as graphical modeling~\cite{kim2009observe,kwon2015unified}.
Modern approaches rely on deep representations (e.g., captured by autoencoders), as in Gaussian classifiers~\cite{SABOKROU2018} and Gaussian Mixtures~\cite{zong2018deep}. In~\cite{hinami2017joint} the authors involve a Kernel Density Estimator (KDE) modeling activations from an auxiliary object detection network.
A recent research trend considers training Generative Adversarial Networks (GANs) on normal samples. However, as such models approximate an implicit density function, they can be queried for new samples but not for likelihood values. Therefore, GAN-based models employ different heuristics for the evaluation of novelty. For instance, in~\cite{schlegl2017unsupervised} a guided latent space search is exploited to infer it, whereas~\cite{ravanbakhsh2017training} directly queries the discriminator for a normality score.
\section{Proposed model}
\label{sec:model}
\begin{figure*}[t!]
\centering
\begin{subfigure}[t]{0.65\textwidth}
\centering
\includegraphics[width=\textwidth]{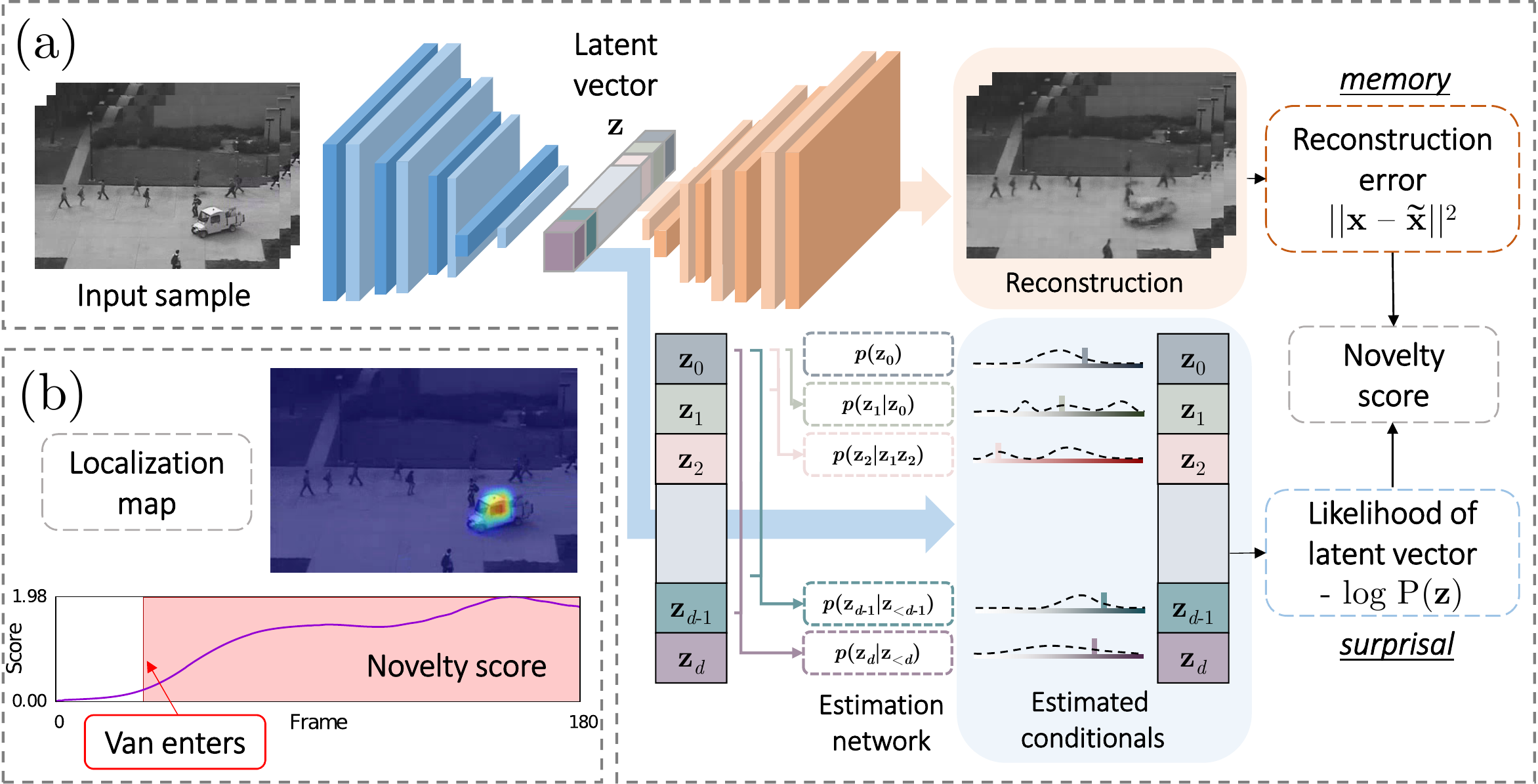}
\caption{}
\label{fig:cover}
\end{subfigure}%
~ 
\begin{subfigure}[t]{0.33\textwidth}
\centering
\includegraphics[width=\textwidth]{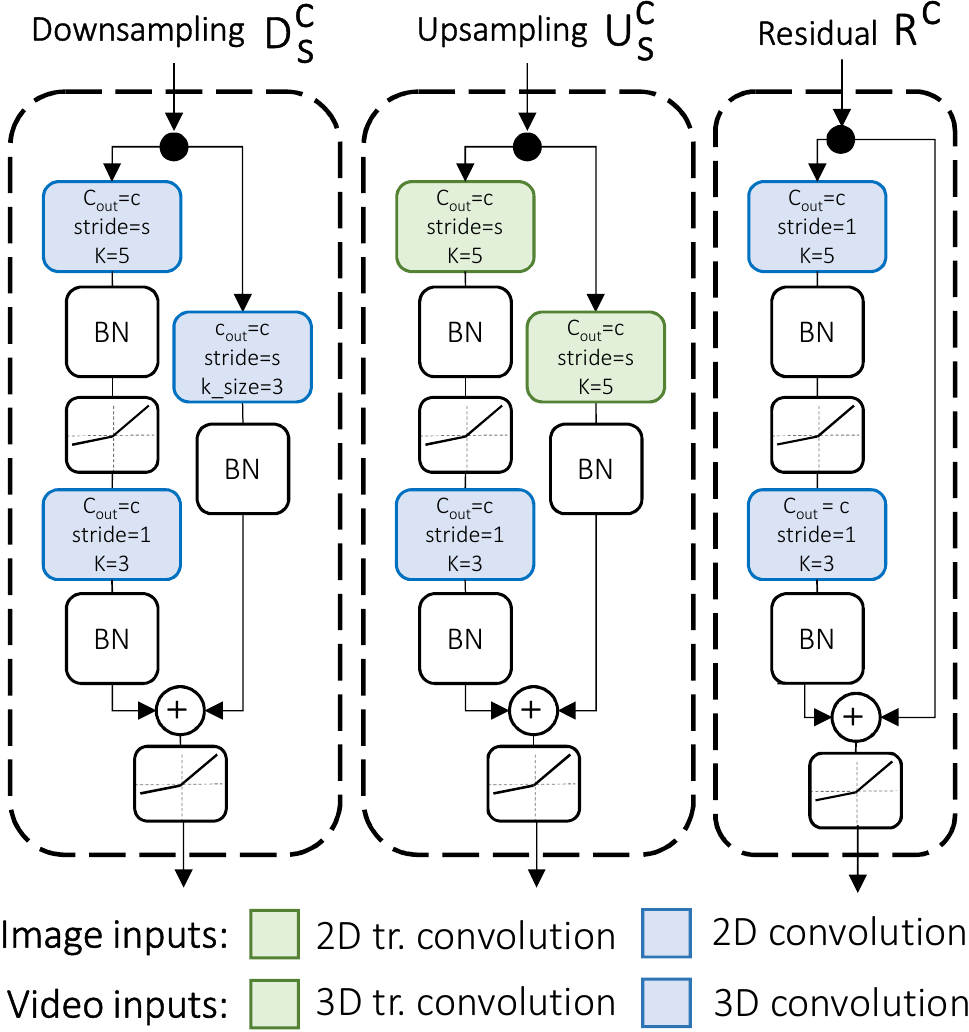}
\caption{}
\label{fig:blocks}
\end{subfigure}
\vspace{-.3cm}
\caption{(i) The proposed novelty detection framework. The overall architecture, depicted in (a), consists of a deep autoencoder and an autoregressive estimation network operating on its latent space. The joint minimization of their respective objective leads to a measure of novelty - (b) - obtained by assessing the remembrance of the model when looking to a new sample, combined with its surprise aroused by causal factors. (ii) Building blocks employed in the autoencoder's architecture.}
\end{figure*}
Maximizing the probability of latent representations is analogous to lowering the surprisal of the model for a normal configuration, defined as the negative log-density of a latent variable instance~\cite{tribus1961thermostatics}. Conversely, remembering capabilities can be evaluated by the reconstruction accuracy of a given sample under its latent representation.\\
We model the aforementioned aspects in a latent variable model setting, where the density function of training samples $p(\textbf{x})$ is modeled through an auxiliary random variable $\textbf{z}$, describing the set of causal factors underlying all observations. By factorizing
\begin{equation}
\label{eq:lte_integral}
p(\textbf{x}) = \int p(\textbf{x}|\textbf{z})p(\textbf{z})d\textbf{z},
\end{equation}
where $p(\textbf{x}|\textbf{z})$ is the conditional likelihood of the observation given a latent representation $\textbf{z}$ with prior distribution $p(\textbf{z})$, we can explicit both the memory and surprisal contribution to novelty.
We approximate the marginalization by means of an inference model responsible for the identification of latent space vector for which the contribution of $p(\textbf{x}|\textbf{z})$ is maximal. Formally, we employ a deep autoencoder, in which the reconstruction error plays the role of the negative logarithm of $p(\textbf{x}|\textbf{z})$, under the hypothesis that $p(\textbf{x}|\textbf{z})=\mathcal{N}(\textbf{x}|\tilde{\textbf{x}},I)$ where $\tilde{\textbf{x}}$ denotes the output reconstruction. Additionally, surprisal is injected in the process by equipping the autoencoder with an auxiliary deep parametric estimator learning the prior distribution $p(\textbf{z})$ of latent vectors, and training it by means of Maximum Likelihood Estimation (MLE). 
Our architecture is therefore composed of three building blocks (Fig.~\ref{fig:cover}): an encoder $f({\bf x};{\bf \theta}_f)$, a decoder $g({\bf z};{\bf \theta}_g)$ and a probabilistic model $h({\bf z};{\bf \theta}_h)$:
\begin{align}
\begin{split}
f({\bf x};{\bf \theta}_f): \mathbb{R}^m& \rightarrow\mathbb{R}^d, \qquad g({\bf z};{\bf \theta}_g):\mathbb{R}^d\rightarrow\mathbb{R}^m, \\
& h({\bf z};{\bf \theta}_h):\mathbb{R}^d\rightarrow[0, 1].
\end{split}
\end{align}
The encoder processes input ${\bf x}$ and maps it into a compressed representation ${\bf z} = f({\bf x};{\bf \theta}_f)$, whereas the decoder provides a reconstructed version of the input $\tilde{{\bf x}} = g({\bf z};{\bf \theta}_g)$. The probabilistic model $h({\bf z};{\bf \theta}_h)$ estimates the density in ${\bf z}$ via an autoregressive process, allowing to avoid the adoption of a specific family of distributions (i.e., Gaussian), potentially unrewarding for the task at hand. On this latter point, please refer to supplementary materials for comparison w.r.t. variational autoencoders \cite{kingma2013auto}.\\
With such modules, at test time, we can assess the two sources of novelty: elements whose observation is poorly explained by the causal factors inducted by normal samples (i.e., high reconstruction error); elements exhibiting good reconstructions whilst showing surprising underlying representations under the learned prior.\\\\
\textbf{Autoregressive density estimation.}
Autoregressive models provide a general formulation for tasks involving sequential predictions, in which each output depends on previous observations~\cite{luc2017predicting,oord2016wavenet}. 
We adopt such a technique to factorize a joint distribution, thus avoiding to define its landscape a priori~\cite{larochelle2011neural, uria2013rnade}. Formally, $p(\textbf{z})$ is factorized as
\begin{equation}
\label{eq:autoregression}
p(\textbf{z})= \prod_{i=1}^d p(z_i|\textbf{z}_{<i}),
\end{equation}
so that estimating $p(\textbf{z})$ reduces to the estimation of each single Conditional Probability Density (CPD) expressed as $p(z_i|\textbf{z}_{<i})$, where the symbol $<$ implies an order over random variables. 
Some prior models obey handcrafted orderings \cite{oord2016pixel,van2016conditional}, whereas others rely on order agnostic training \cite{uria2014deep,germain2015made}. Nevertheless, it is still not clear how to estimate the proper order for a given set of variables. In our model, this issue is directly tackled by the optimization. Indeed, since we perform autoregression on learned latent representations, the MLE objective encourages the autoencoder to impose over them a pre-defined causal structure. Empirical evidence of this phenomenon is given in the supplementary material.\\
From a technical perspective, the estimator $h({\bf z};{\bf \theta}_h)$ outputs parameters for $d$ distributions $p(z_i|{\bf z}_{<i})$. In our implementation, each CPD is modeled as a multinomial over B=100 quantization bins. To ensure a conditional estimate of each underlying density, we design proper layers guaranteeing that the CPD of each symbol $z_i$ is computed from inputs $\{z_1, \ldots, z_{i-1}\}$ only.\\\\
\textbf{Objective and connection with differential entropy.}
The three components $f$, $g$ and $h$ are jointly trained to minimize $\mathcal{L} \equiv \mathcal{L}(\theta_f, \theta_g, \theta_h)$ as follows:
\begin{align}
\begin{split}
\label{eq:loss}
\mathcal{L} &= \mathcal{L}_{\text{REC}}(\theta_f, \theta_g) + \lambda \mathcal{L}_{\text{LLK}}(\theta_f, \theta_h) \\ 
&=\mathbb{E}_{\textbf{x}}\bigg[{\underbrace{||\textbf{x} - \tilde{\textbf{x}}||^2}_{\text{reconstruction term}}} - \lambda\underbrace{\log(h(\textbf{z}; \theta_h))}_{\text{log-likelihood term}}\bigg],
\end{split}
\end{align}
where $\lambda$ is a hyper-parameter controlling the weight of the $\mathcal{L}_{\text{LLK}}$ term. It is worth noting that it is possible to express the log-likelihood term as
\begin{align}
\begin{split}
\label{eq:kl_and_entropy}
&\mathbb{E}_{\textbf{z} \sim p^*(\textbf{z};\theta_f)}\big[-\log h(\textbf{z}; \theta_h)\big]\\
=& \ \mathbb{E}_{\textbf{z} \sim p^*(\textbf{z};\theta_f)}\big[ - \log h(\textbf{z}; \theta_h) + \log p^*(\textbf{z};\theta_f)-\log p^*(\textbf{z}; \theta_f)\big]\\
=& \ D_{\mathrm {KL} }(p^*(\textbf{z}; \theta_f)\ \| \ {h(\textbf{z}; \theta_h))} + \mathbb{H}[p^*(\textbf{z}; \theta_f)],
\end{split}
\end{align}
\begin{figure}[t]
\centering
\includegraphics[width=0.9\columnwidth]{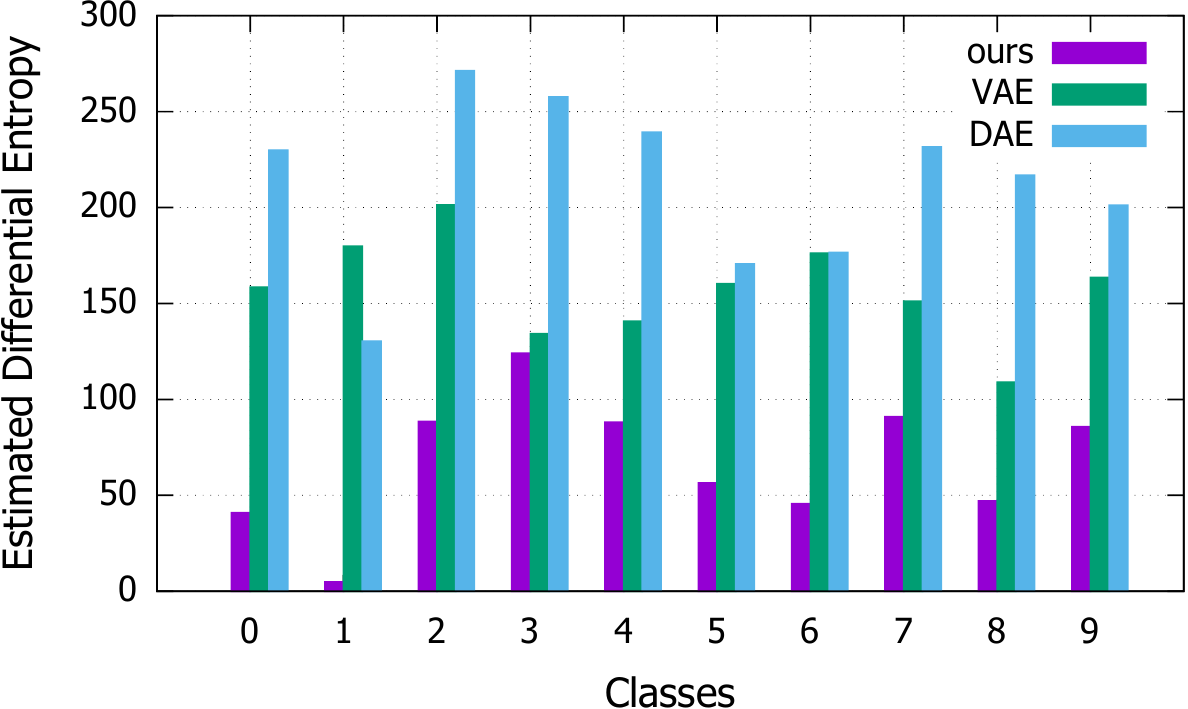}
\caption{Estimated differential entropies delivered on each MNIST class in the presence of different regularization strategies: our, divergence w.r.t a Gaussian prior (VAE) and input perturbation (DAE). For each class, the estimate is computed on the training samples' hidden representations, whose distribution are fit utilizing a Gaussian KDE in a 3D-space. All models being equal, ours exhibits lower entropies on all classes.}
\label{fig:mnist_entropy}
\vspace{-.3cm}
\end{figure}
\begin{figure*}[t]
\begin{center}
\resizebox{\textwidth}{!}{%
\begin{tabular}{cc}
\includegraphics[width=0.45\textwidth]{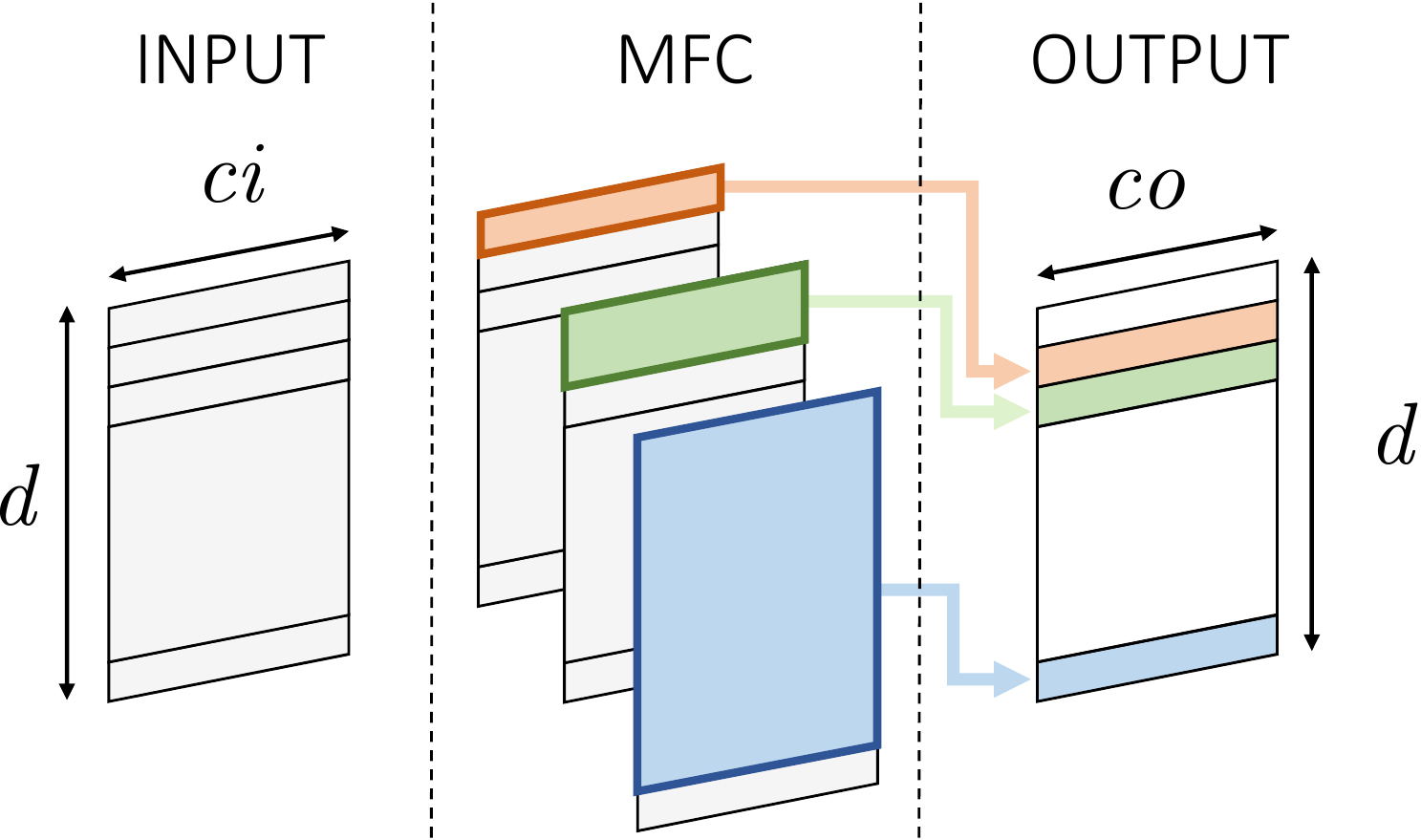}&
\includegraphics[width=0.62\textwidth]{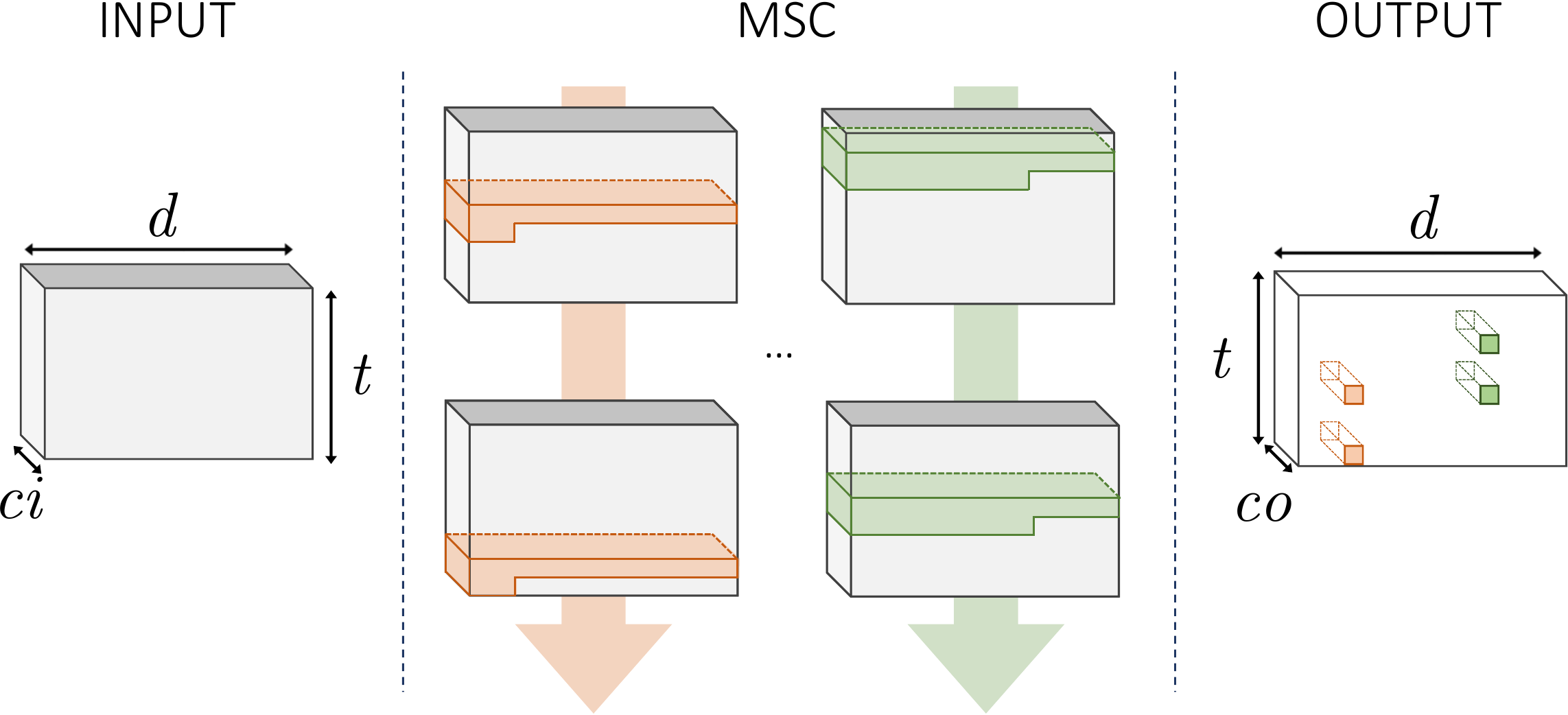}\\
(a)&(b)
\end{tabular}}
\end{center}
\caption{Proposed autoregressive layers, namely the Masked Fully Connection (a, Eq.~\ref{eq:mfc}) and the Masked Stacked Convolution (b, Eq.~\ref{eq:msc}). For both layers, we represent type A structure. Different kernel colors represent different parametrizations.}
\label{fig:custom_layers}
\vspace{-.3cm}
\end{figure*}
where $p^*(\textbf{z}; \theta_f)$ denotes the true distribution of the codes produced by the encoder, and is therefore parametrized by $\theta_f$. This reformulation of the MLE objective yields meaningful insights about the entities involved in the optimization. On the one hand, the Kullback-Leibler divergence ensures that the information gap between our parametric model $h$ and the true distribution $p^*$ is small. On the other hand, this framework leads to the minimization of the differential entropy of the distribution underlying the codes produced by the encoder $f$. Such constraint constitutes a crucial point when learning normality. 
Intuitively, if we think about the encoder as a source emitting symbols (namely, the latent representations), its desired behavior, when modeling normal aspects in the data, should converge to a `boring' process characterized by an intrinsic low entropy, since surprising and novel events are unlikely to arise during the training phase. Accordingly, among all the possible settings of the hidden representations, the objective begs the encoder to exhibit a low differential entropy, leading to the extraction of features that are easily predictable, therefore common and recurrent within the training set. This kind of features is indeed the most useful to distinguish novel samples from the normal ones, making our proposal a suitable regularizer in the anomaly detection setting.\\
We report empirical evidence of the decreasing differential entropy in Fig.~\ref{fig:mnist_entropy}, that compares the behavior of the same model under different regularization strategies.
\subsection{Architectural Components}
\label{sec:architectural_components}
\textbf{Autoencoder blocks.} 
Encoder and decoder are respectively composed by downsampling and upsampling residual blocks depicted in Fig.~\ref{fig:blocks}. The encoder ends with fully connected (FC) layers. When dealing with video inputs, we employ \emph{causal} 3D convolutions~\cite{BaiTCN2018} within the encoder (i.e., only accessing information from previous time-steps). Moreover, at the end of the encoder, we employ a temporally-shared full connection (TFC, namely a linear projection sharing parameters across the time axis on the input feature maps) resulting in a temporal series of feature vectors. This way, the encoding procedure does not shuffle information across time-steps, ensuring temporal ordering.\\\\
\textbf{Autoregressive layers.}
To guarantee the autoregressive nature of each output CPD, we need to ensure proper connectivity patterns in each layer of the estimator $h$. 
Moreover, since latent representations exhibit different shapes depending on the input nature (image or video), we propose two different solutions.\\
When dealing with images, the encoder provides feature vectors with dimensionality $d$. The autoregressive estimator is composed by stacking multiple Masked Fully Connections (MFC, Fig.~\ref{fig:custom_layers}-(a)).
Formally, it computes output feature map ${\bf o} \in \mathbb{R}^{d\times co}$ (where $co$ is the number of output channels) given the input ${\bf h} \in \mathbb{R}^{d\times ci}$ (assuming $ci=1$ at the input layer). The connection between the input element ${\bf h}_i^k$ in position $i$, channel $k$ and the output element ${\bf o}_j^l$ 
is parametrized by 
\begin{align}
\label{eq:mfc}
\begin{cases}
w_{i, j}^{k, l} &\text{if $i < j$}\\[3pt]
\begin{cases}
w_{i, j}^{k, l} \quad &\text{if type = B}\\
0 \quad &\text{if type = A}\\
\end{cases} \quad &\text{if $i = j$}\\
0 &\text{if $i > j$.}\\
\end{cases}
\end{align}
Type A forces a strict dependence on previous elements (and is employed only as the first estimator layer), whereas type B masks only succeeding elements.
Assuming each CPD modeled as a multinomial, the output of the last autoregressive layer (in $\mathbb{R}^{d \times B}$) provides probability estimates for the $B$ bins that compose the space quantization.\\
On the other hand, the compressed representation of video clips has dimensionality $t\times d$, being $t$ the number of temporal time-steps and $d$ the length of the code. Accordingly, the estimation network is designed to capture two-dimensional patterns within observed elements of the code.
However, naively plugging 2D convolutional layers would assume translation invariance on both axes of the input map, whereas, due to the way the compressed representation is built, this assumption is only correct along the temporal axis.
To cope with this, we apply $d$ different convolutional kernels along the code axis, allowing the observation of the whole feature vector in the previous time-step as well as a portion of the current one. Every convolution is free to stride along the time axis and captures temporal patterns.
In such operation, named Masked Stacked Convolution (MSC, Fig.~\ref{fig:custom_layers}-(b)), the $i$-th convolution is equipped with a kernel ${\bf w}^{(i)} \in \mathbb{R}^{3\times d}$ kernel, that gets multiplied by the binary mask ${\bf M}^{(i)}$, defined as 
\begin{equation}
\label{eq:msc}
m_{j,k}^{(i)}\in {\bf M}^{(i)}=
\begin{cases}
1 \quad \text{if $j=0$}\\
1 \quad \text{if $j=1$ and $k<i$ and type=A}\\
1 \quad \text{if $j=1$ and $k\leq i$ and type=B}\\
0 \quad \text{otherwise,}
\end{cases}
\end{equation}
where $j$ indexes the temporal axis and $k$ the code axis.\\
Every single convolution yields a column vector, as a result of its stride along time. The set of column vectors resulting from the application of the $d$ convolutions to the input tensor ${\bf h} \in \mathbb{R}^{t\times d \times ci}$ are horizontally stacked to build the output tensor ${\bf o} \in \mathbb{R}^{t\times d \times co}$, as follows:
\begin{equation}
\label{eq:conv_stack}
{\bf o}=\concat\limits_{i=1}^{d} [({\bf M}^{(i)}\odot {\bf w}^{(i)})\ast {\bf h}],
\end{equation}
where $||$ represents the horizontal concatenation operation.\\
\begin{table*}[t]
\begin{center}
\resizebox{0.9\textwidth}{!}{%
\begin{tabular}{cCCCCCCCcCCCCCCC}
\toprule
&\multicolumn{7}{c}{MNIST}&&\multicolumn{7}{c}{CIFAR10}\\\cmidrule{2-8}\cmidrule{10-16}
& OC SVM & KDE & DAE & VAE & Pix CNN & GAN & ours && OC SVM & KDE & DAE & VAE & Pix CNN & GAN & ours\\\cmidrule{2-8}\cmidrule{10-16}
0&0.988&0.885&0.991&0.998&0.531&0.926&0.993&&0.630&0.658&0.718&0.688&0.788&0.708&0.735\\
1&0.999&0.996&0.999&0.999&0.995&0.995&0.999&&0.440&0.520&0.401&0.403&0.428&0.458&0.580\\
2&0.902&0.710&0.891&0.962&0.476&0.805&0.959&&0.649&0.657&0.685&0.679&0.617&0.664&0.690\\
3&0.950&0.693&0.935&0.947&0.517&0.818&0.966&&0.487&0.497&0.556&0.528&0.574&0.510&0.542\\
4&0.955&0.844&0.921&0.965&0.739&0.823&0.956&&0.735&0.727&0.740&0.748&0.511&0.722&0.761\\
5&0.968&0.776&0.937&0.963&0.542&0.803&0.964&&0.500&0.496&0.547&0.519&0.571&0.505&0.546\\
6&0.978&0.861&0.981&0.995&0.592&0.890&0.994&&0.725&0.758&0.642&0.695&0.422&0.707&0.751\\
7&0.965&0.884&0.964&0.974&0.789&0.898&0.980&&0.533&0.564&0.497&0.500&0.454&0.471&0.535\\
8&0.853&0.669&0.841&0.905&0.340&0.817&0.953&&0.649&0.680&0.724&0.700&0.715&0.713&0.717\\
9&0.955&0.825&0.960&0.978&0.662&0.887&0.981&&0.508&0.540&0.389&0.398&0.426&0.458&0.548\\\cmidrule{2-8}\cmidrule{10-16}
avg&0.951&0.814&0.942&0.969&0.618&0.866&\textbf{0.975}&&0.586&0.610&0.590&0.586&0.551&0.592&\textbf{0.641}\\\bottomrule
\end{tabular}}
\end{center}
\caption{AUROC results for novelty detection on MNIST and CIFAR10. Each row represents a different class on which baselines and our model are trained.}
\label{tab:oc_mnist_cifar}
\vspace{-.3cm}
\end{table*}
\section{Experiments\protect\footnote{Code to reproduce results in this section is released at \url{https://github.com/aimagelab/novelty-detection}.}}
\label{sec:experiments}
We test our solution in three different settings: images, videos, and cognitive data. In all experiments the novelty assessment on the $i$-th example is carried out by summing the reconstruction term (REC$_i$) and the log-likelihood term (LLK$_i$) in Eq.~\ref{eq:loss} in a single novelty score $NS_i$:
\begin{equation}
\label{eq:loss_fusion}
NS_i = norm_S(REC_i) + norm_S(LLK_i).
\end{equation}
Individual scores are normalized using a reference set of examples $S$ (different for every experiment),
\begin{equation}
norm_S(L_i) = \frac{L_i - \max_{j\in S} L_j}{\max_{j\in S} L_j - \min_{j\in S} L_j}.
\end{equation}
Further implementation details and architectural hyperparameters are in the supplementary material.
\subsection{One-class novelty detection on images}
\label{sec:one_class_novelty_detection}
To assess the model's performances in one class settings, we train it on each class of either MNIST or CIFAR-10 separately. In the test phase, we present the corresponding test set, which is composed of 10000 examples of all classes, and expect our model to assign a lower novelty score to images sharing the label with training samples. We use standard train/test splits, and isolate 10\% of training samples for validation purposes, and employ it as the normalization set ($S$ in Eq.~\ref{eq:loss_fusion}) for the computation of the novelty score.\\
As for the baselines, we consider the following:
\begin{itemize}
\setlength\itemsep{-0.3em}
\item[-]standard methods such as OC-SVM~\cite{scholkopf2000support} and Kernel Density Estimator (KDE), employed out of features extracted by PCA-whitening;
\item[-]a denoising autoencoder (DAE) sharing the same architecture as our proposal, but defective of the density estimation module. The reconstruction error is employed as a measure of normality vs. novelty;
\item[-]a variational autoencoder (VAE)~\cite{kingma2013auto}, also sharing the same architecture as our model, in which the Evidence Lower Bound (ELBO) is employed as the score;
\item[-]Pix-CNN~\cite{van2016conditional}, modeling the density by applying autoregression directly in the image space;
\item[-]the GAN-based approach illustrated in~\cite{schlegl2017unsupervised}.
\end{itemize}
We report the comparison in Tab.~\ref{tab:oc_mnist_cifar} in which performances are measured by the Area Under Receiver Operating Characteristic (AUROC), which is the standard metric for the task.
As the table shows, our proposal outperforms all baselines in both settings.\\\\
Considering MNIST, most methods perform favorably.
Notably, Pix-CNN fails in modeling distributions for all digits but one, possibly due to the complexity of modeling densities directly on pixel space and following a fixed autoregression order. Such poor test performances are registered despite good quality samples that we observed during training: indeed, the weak correlation between sample quality and test log-likelihood of the model has been motivated in~\cite{theis2015note}.
Surprisingly, OC-SVM outperforms most deep learning based models in this setting.\\
On the contrary, CIFAR10 represents a much more significant challenge, as testified by the low performances of most models, possibly due to the poor image resolution and visual clutter between classes. Specifically, we observe that our proposal is the only model outperforming a simple KDE baseline; however, this finding should be put into perspective by considering the nature of non-parametric estimators. Indeed, non-parametric models are allowed to access the whole training set for the evaluation of each sample. Consequently, despite they benefit large sample sets in terms of density modeling, they lead into an unfeasible inference as the dataset grows in size.
\begin{figure}[b]
\centering
\resizebox{\columnwidth}{!}{
\begin{tabular}{cc}
\includegraphics{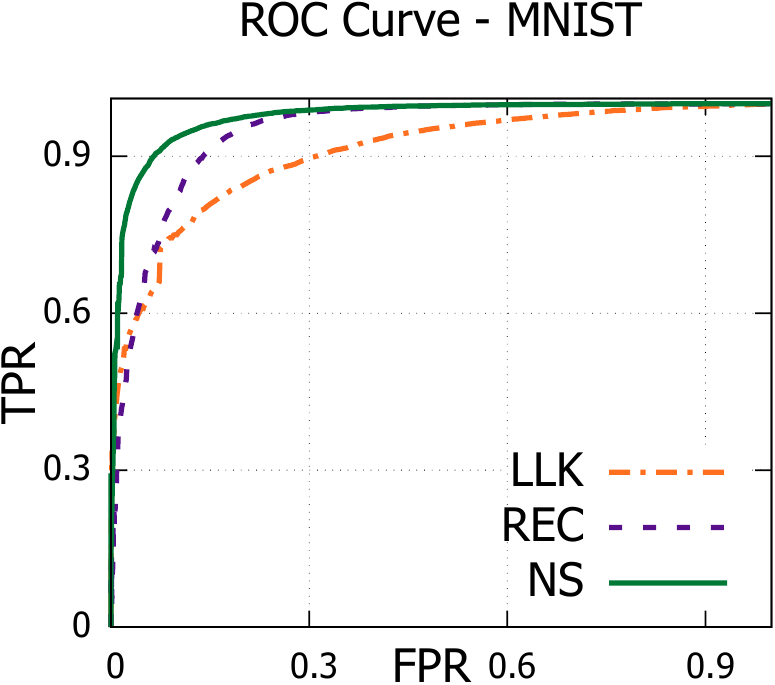}&
\includegraphics{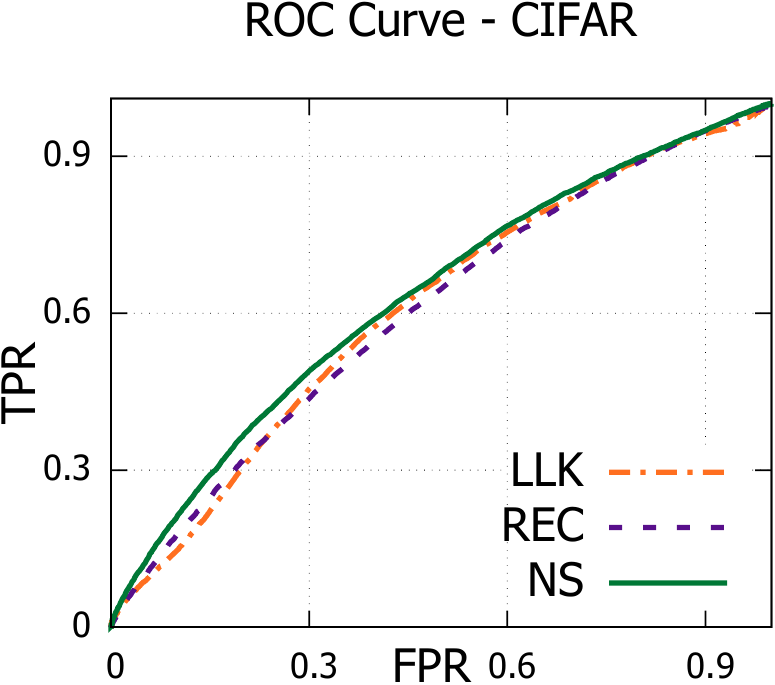}
\end{tabular}}
\caption{ROC curves delivered by different scoring strategies on MNIST and CIFAR-10 test sets. Each curve is an interpolation over the ten classes.}
\label{fig:loss_fusion}
\vspace{-.3cm}
\end{figure}
\\The possible reasons behind the difference in performance w.r.t. DAE are twofold. Firstly, DAE can recognize novel samples solely based on the reconstruction error, hence relying on its memorization capabilities, whereas our proposal also considers the likelihood of their representations under the learned prior, thus exploiting surprisal as well. Secondly, by minimizing the differential entropy of the latent distribution, our proposal increases the discriminative capability of the reconstruction. Intuitively, this last statement can be motivated observing that novelty samples are forced to reside in a high probability region of the latent space, the latter bounded to solely capture unsurprising factors of variation arising from the training set. 
On the other hand, the gap w.r.t. VAE suggests that, for the task at hand, a more flexible autoregressive prior should be preferred over the isotropic multivariate Gaussian. 
On this last point, VAE seeks representations whose average surprisal converges to a fixed and expected value (i.e., the differential entropy of its prior), whereas our solution minimizes such quantity within its MLE objective. This flexibility allows modulating the richness of the latent representation vs. the reconstructing capability of the model. On the contrary, in VAEs, the fixed prior acts as a blind regularizer, potentially leading to over-smooth representations; this aspect is also appreciable when sampling from the model as shown in the supplementary material.\\\\
Fig.~\ref{fig:loss_fusion} reports an ablation study questioning the loss functions aggregation presented in Eq.~\ref{eq:loss_fusion}. The figure illustrates ROC curves under three different novelty scores: i) the log-likelihood term, ii) the reconstruction term, and iii) the proposed scheme that accounts for both. As highlighted in the picture, accounting for both memorization and surprisal aspects is advantageous in each dataset.
Please refer to the supplementary material for additional evidence.
\subsection{Video anomaly detection}
\label{sec:video_anomaly_detection}
\begin{figure*}
\centering
\begin{minipage}{0.4\textwidth}
\resizebox{\textwidth}{!}{
\begin{tabular}{rcc}
\toprule
& UCSD Ped2 & ShanghaiTech\\\midrule
MPPCA~\cite{kim2009observe}&0.693&-\\
MPPC+SFA~\cite{mahadevan2010anomaly}&0.613&-\\
MDT~\cite{mahadevan2010anomaly}&0.829&-\\
ConvAE~\cite{hasan2016learning}&0.850&0.609\\
ConvLSTM-AE~\cite{luo2017remembering}&0.881&-\\
Unmasking~\cite{ionescu2017unmasking}&0.822&-\\
Hinami~\etal~\cite{hinami2017joint}&0.922&-\\
TSC~\cite{luo2017revisit}&0.910&0.679\\
Stacked RNN~\cite{luo2017revisit}&0.922&0.680\\
FFP~\cite{liu2017future}&0.935&-\\
FFP+MC~\cite{liu2017future}&\textbf{0.954}&\textbf{0.728}\\\midrule
Ours &\textbf{0.954}&\textbf{0.725}\\\bottomrule
\end{tabular}}
\end{minipage}
\begin{minipage}{0.55\textwidth}
\resizebox{\textwidth}{!}{
\begin{tabular}{cc}
\includegraphics[width=0.5\textwidth]{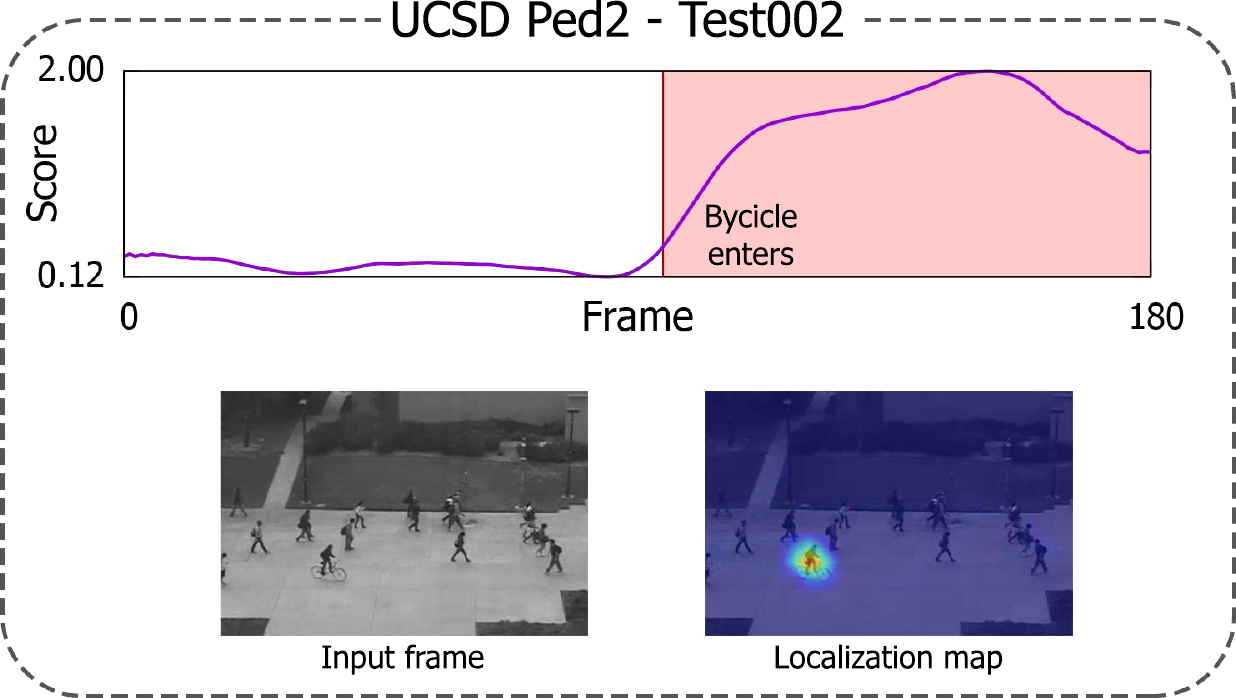}&
\includegraphics[width=0.5\textwidth]{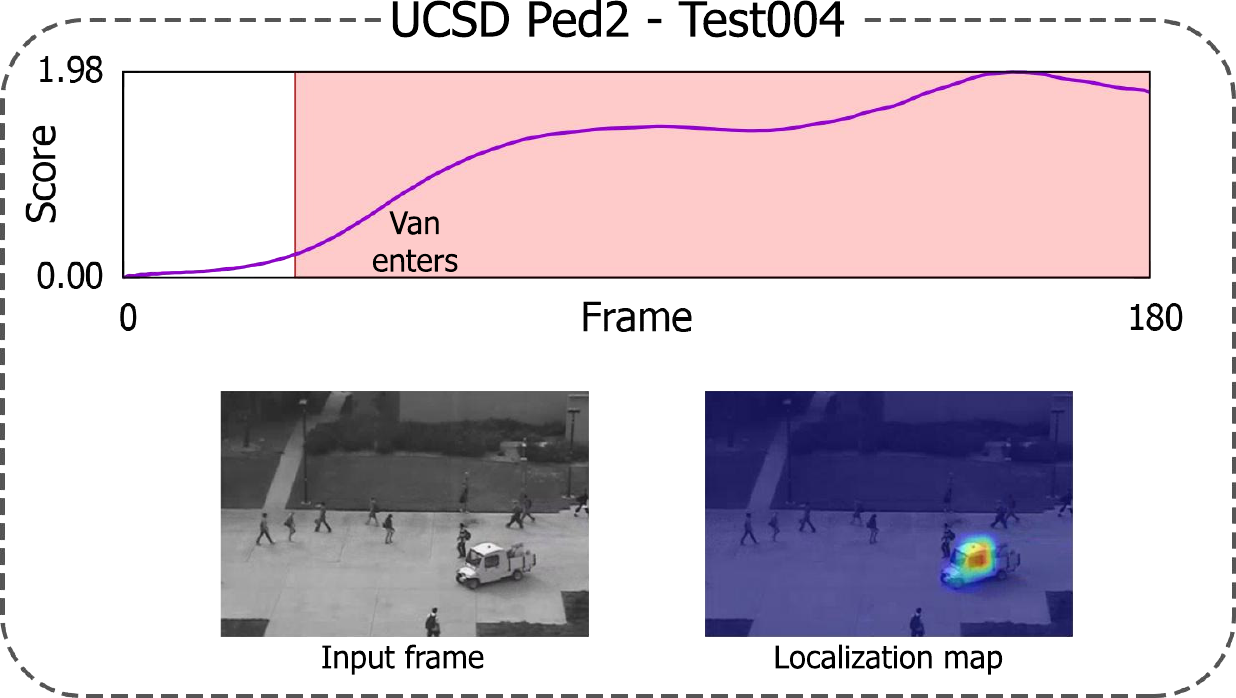}\\
\includegraphics[width=0.5\textwidth]{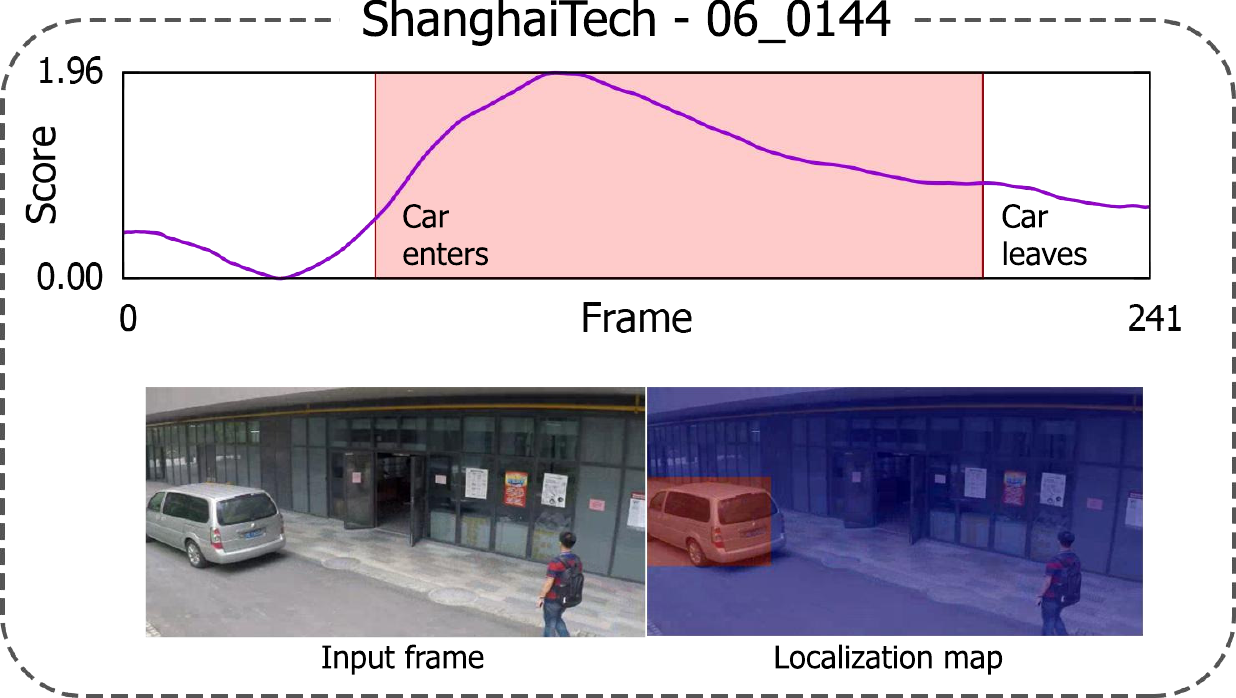}&
\includegraphics[width=0.5\textwidth]{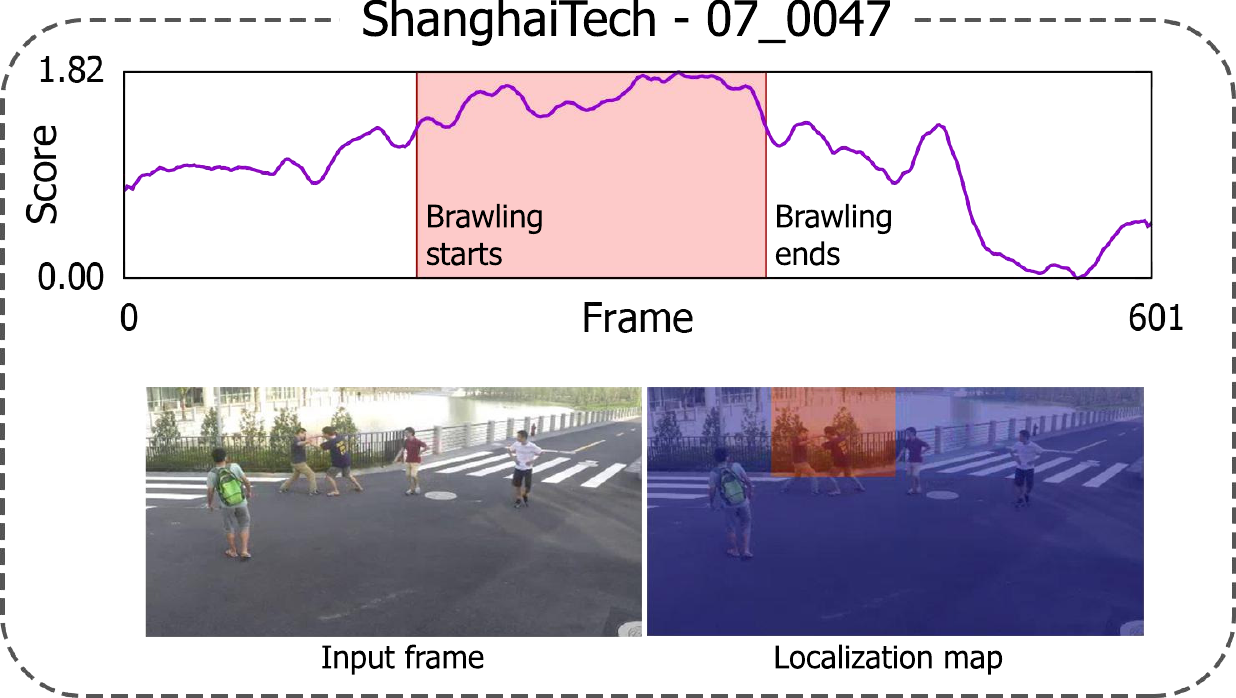}
\end{tabular}}
\end{minipage}
\caption{On the left, AUROC performances of our model w.r.t. state-of-the-art competitors. On the right, novelty scores and localizations maps for samples drawn from UCSD Ped2 and ShanghaiTech. For each example, we report the trend of the assessed score, highlighting with a different color the time range in which an anomalous subject comes into the scene.}
\label{fig:video_anomaly_detection}
\vspace{-.3cm}
\end{figure*}
In video surveillance contexts, novelty is often considered in terms of abnormal human behavior. Thus, we evaluate our proposal against state-of-the-art anomaly detection models. For this purpose, we considered two standard benchmarks in literature, namely UCSD Ped2~\cite{chan2008ucsd} and ShanghaiTech~\cite{luo2017revisit}. Despite the differences in the number of videos and their resolution, they both contain anomalies that typically arise in surveillance scenarios (e.g., vehicles in pedestrian walkways, pick-pocketing, brawling).
For UCSD Ped, we preprocessed input clips of 16 frames to extract smaller patches (we refer to supplementary materials for details) and perturbed such inputs with random Gaussian noise with $\sigma=0.025$. We compute the novelty score of each input clip as the mean novelty score among all patches. Concerning ShanghaiTech, we removed the dependency on the scenario by estimating the foreground for each frame of a clip with a standard MOG-based approach and removing the background. We fed the model with 16-frames clips, but ground-truth anomalies are labeled at frame level. In order to recover the novelty score of each frame, we compute the mean score of all clips in which it appears. We then merge the two terms of the loss function following the same strategy illustrated in Eq.~\ref{eq:loss_fusion}, computing however normalization coefficients in a per-sequence basis, following the standard approach in the anomaly detection literature. The scores for each sequence are then concatenated to compute the overall AUROC of the model. 
Additionally, we envision localization strategies for both datasets. To this aim, for UCSD, we denote a patch exhibiting the highest novelty score in a frame as anomalous. Differently, in ShanghaiTech, we adopt a sliding-window approach~\cite{zeiler2014visualizing}: as expected, when occluding the source of the anomaly with a rectangular patch, the novelty score drops significantly.\\
Fig.~\ref{fig:video_anomaly_detection} reports results in comparison with prior works, along with qualitative assessments regarding the novelty score and localization capabilities. Despite a more general formulation, our proposal scores on-par with the current state-of-the-art solutions specifically designed for video applications and taking advantage of optical flow estimation and motion constraints. Indeed, in the absence of such hypotheses (FFP entry in Fig.~\ref{fig:video_anomaly_detection}), our method outperforms future frame prediction on UCSD Ped2.
\subsection{Model Analysis}
\label{sec:model_analysis}
\begin{figure}[t]
\begin{minipage}{0.5\columnwidth}
\centering
\includegraphics[width=\textwidth]{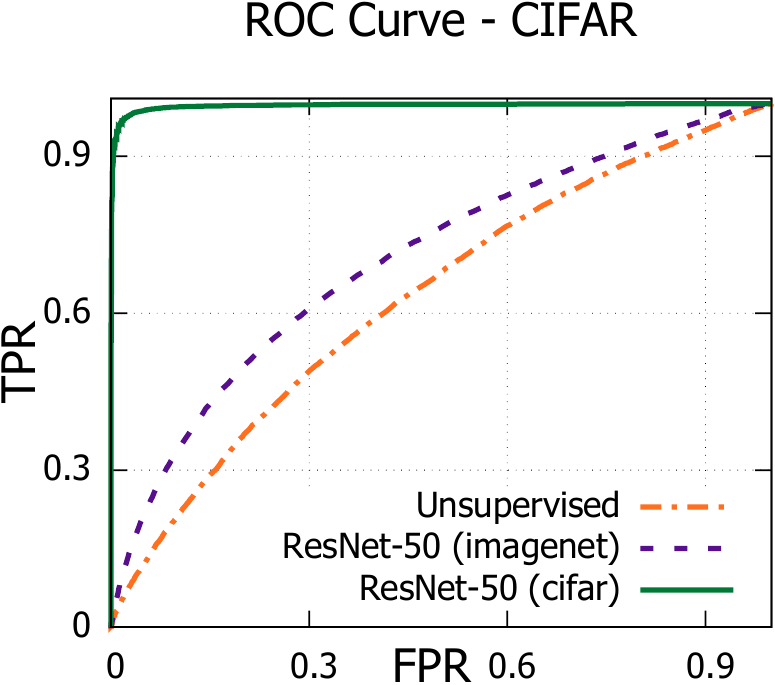}\\
(a)
\end{minipage}
\begin{minipage}{0.47\columnwidth}
\resizebox{\textwidth}{!}{
\begin{tabular}{lc}
\hline
\multicolumn{2}{c}{CIFAR-10}\\\hline
LSTM$_{[100]}$ & 0.623\\
LSTM$_{[32,32,32,32,100]}$ & 0.622\\
MFC$_{[100]}$ & 0.625\\
MFC$_{[32,32,32,32,100]}$& \textbf{0.641}\\\hline\hline
\multicolumn{2}{c}{UCSD Ped2}\\\hline
LSTM$_{[100]}$ & 0.849\\
LSTM$_{[4,4,4,4,100]}$ & 0.845 \\
MSC$_{[100]}$ & 0.849\\
MSC$_{[4,4,4,4,100]}$& \textbf{0.954}\\\hline
\end{tabular}}\\
\\
\vspace{-.26cm}
\\
\centering (b)
\end{minipage}
\caption{(a) CIFAR-10 ROC curves with semantic input vectors. Each curve is an interpolation among the ten classes. (b) Comparison of different architectures for the autoregressive density estimation in feature space. We indicate with LSTM$_{[F_1,F_2,\dots,F_N]}$ - same goes for MFC and MSC - the output shape for each of the $N$ layers composing the estimator. Results are reported in terms of test AUROC.}
\label{fig:ablation}
\vspace{-.3cm}
\end{figure}

\textbf{CIFAR-10 with semantic features.}
We investigate the behavior of our model in the presence of different assumptions regarding the expected nature of novel samples. We expect that, as the correctness of such assumptions increases, novelty detection performances will scale accordingly. Such a trait is particularly desirable for applications in which prior beliefs about novel examples can be envisioned.
To this end, we leverage the CIFAR-10 benchmark described in Sec.~\ref{sec:one_class_novelty_detection} and change the type of information provided as input. Specifically, instead of raw images, we feed our model with semantic representations extracted by ResNet-50~\cite{he2016deep}, either pre-trained on Imagenet (i.e., assume semantic novelty) or CIFAR-10 itself (i.e., assume data-specific novelty). The two models achieved respectively 79.26 and 95.4 top-1 classification accuracies on the respective test sets.
Even though this procedure is to be considered unfair in novelty detection, it serves as a sanity check delivering the upper-bound performances our model can achieve when applied to even better features. To deal with dense inputs, we employ a fully connected autoencoder and MFC layers within the estimation network.\\
Fig.~\ref{fig:ablation}-(a) illustrates the resulting ROC curves, where semantic descriptors improve AUROC w.r.t. raw image inputs (entry \quotes{Unsupervised}). Such results suggest that our model profitably takes advantage of the separation between normal and abnormal input representations and scales accordingly, even up to optimal performances for the task under consideration. Nevertheless, it is interesting to note how different degrees of supervision deliver significantly different performances. As expected, dataset-specific supervision increases the AUROC from 0.64 up to 0.99 (a perfect score). Surprisingly, semantic feature vectors trained on Imagenet (which contains all CIFAR classes) provide a much lower boost, yielding an AUROC of 0.72. 
Such result suggests that, even in the rare cases where the semantic of novelty can be known in advance, its contribution has a limited impact in modeling the normality, mostly because novelty can depend on other cues (e.g., low-level statistics).\\\\
\textbf{Autoregression via recurrent layers.}
To measure the contribution of the proposed MFC and MSC layers described in Sec.~\ref{sec:model}, we test on CIFAR-10 and UCSD Ped2, alternative solutions for the autoregressive density estimator.
Specifically, we investigate recurrent networks, as they represent the most natural alternative featuring autoregressive properties.
We benchmark the proposed building blocks against an estimator composed of LSTM layers, which is designed to sequentially observe latent symbols $\textbf{z}_{<i}$ and output the CPD of $z_i$ as the hidden state of the last layer. We test MFC, MSC and LSTM in single-layer and multi-layer settings, and report all outcomes in Fig.~\ref{fig:ablation}-(b).
\\It emerges that, even though our solutions perform similarly to the recurrent baseline when employed in a shallow setting, they significantly take advantage of their depth when stacked in consecutive layers. MFC and MSC, indeed, employ disentangled parametrizations for each output CPD. This property is equivalent to the adoption of a specialized estimator network for each $z_i$, thus increasing the proficiency in modeling the density of its designated CPD. On the contrary, LSTM networks embed all the history (i.e., the observed symbols) in their memory cells, but manipulate each input of the sequence through the same weight matrices. In such a regime, the recurrent module needs to learn parameters shared among symbols, losing specialization and eroding its modeling capabilities.
\subsection{Novelty in cognitive temporal processes}
\begin{figure}[t]
\centering
\resizebox{\columnwidth}{!}{
\begin{tabular}{cc}
\multirow{2}{*}{\includegraphics[width=0.7\columnwidth]{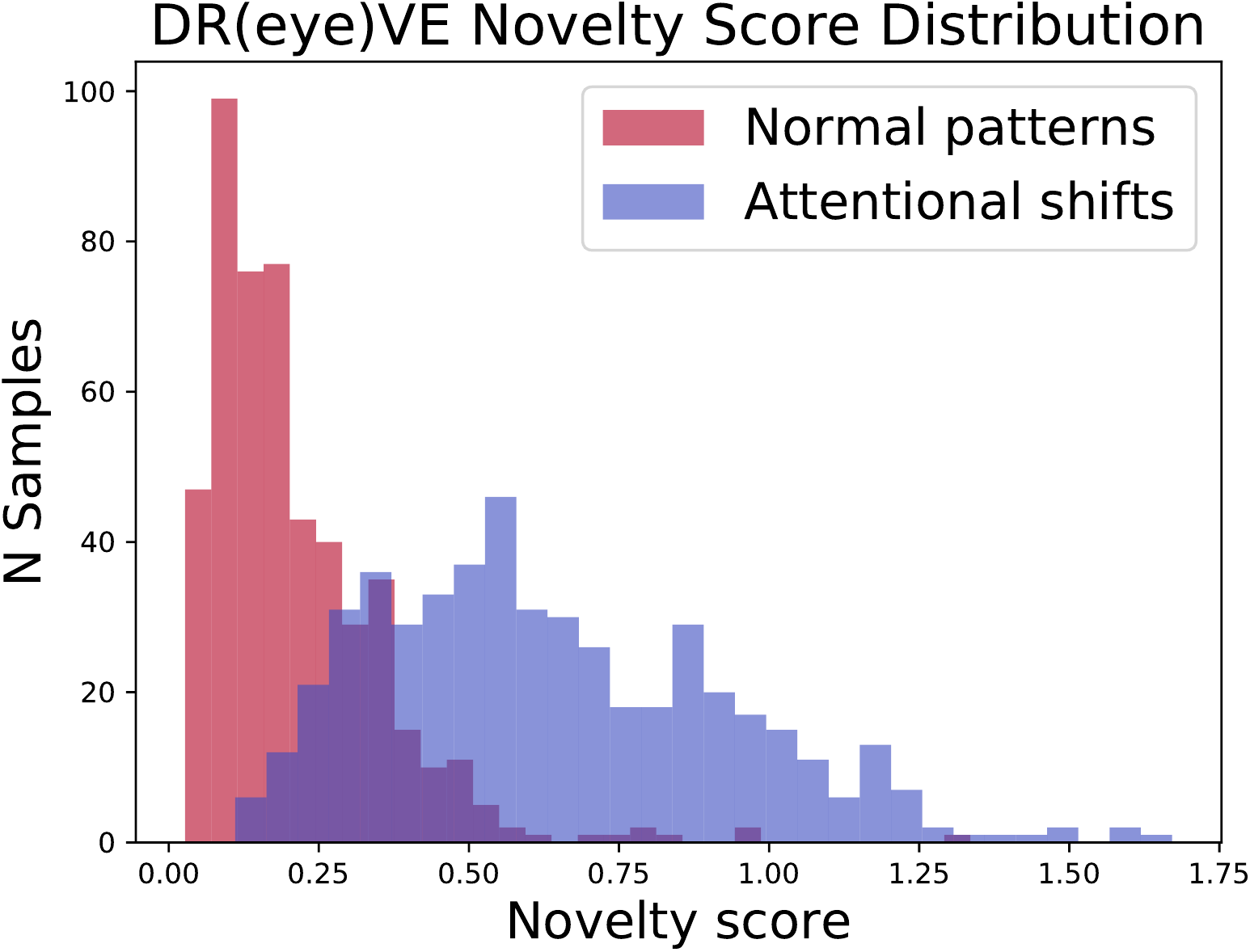}}\\
&\includegraphics[width=0.35\columnwidth]{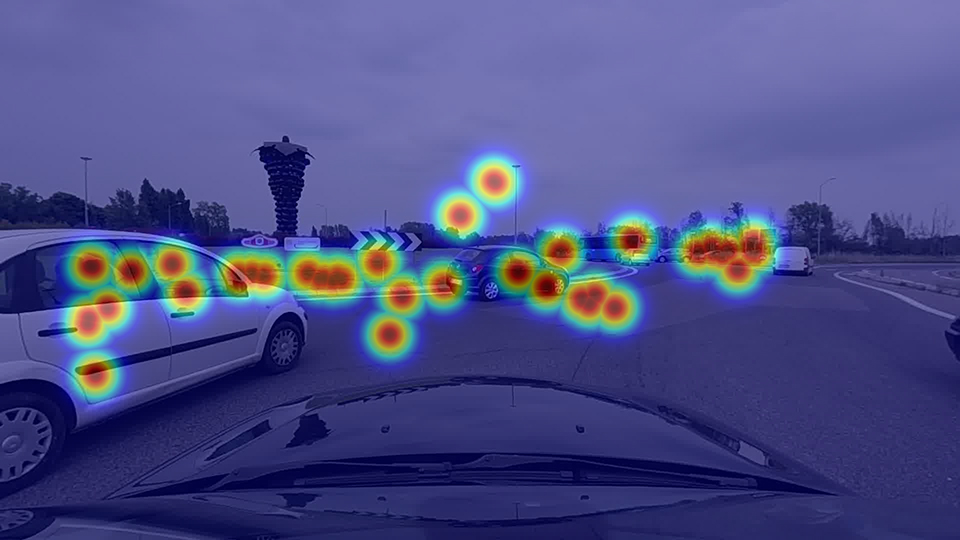}\\
&\includegraphics[width=0.35\columnwidth]{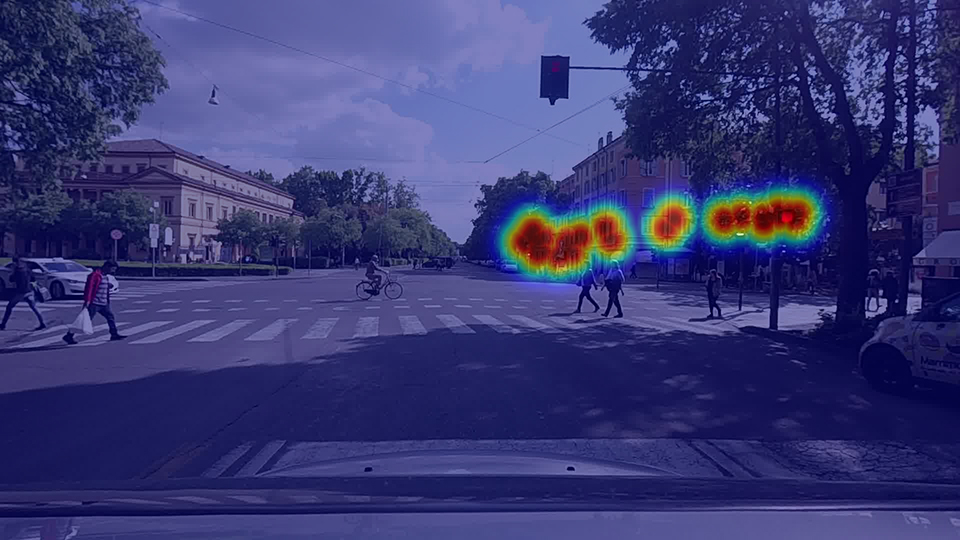}
\end{tabular}}
\vspace{.3cm}
\caption{Left, the distribution of novelty scores assigned to normal patterns against attentional shifts labeled within the DR(eye)VE dataset. Right, DR(eye)VE clips yielding the highest novelty score (i.e., clips in which the attentional pattern shifts from the expected behavior). Interestingly, they depict some peculiar situations such as waiting for the traffic light or approaching a roundabout.}
\label{fig:dreyeve}
\vspace{-.3cm}
\end{figure}

As a potential application of our proposal, we investigate its capability in modeling human attentional behavior. To this end, we employ the DR(eye)VE dataset~\cite{dreyeve2018}, introduced for the prediction of focus of attention in driving contexts. It features 74 driving videos where frame-wise fixation maps are provided, highlighting the region of the scene attended by the driver. In order to capture the dynamics of attentional patterns, we purposely discard the visual content of the scene and optimize our model on clips of fixation maps, randomly extracted from the training set. After training, we rely on the novelty score of each clip as a proxy for the uncommonness of an attentional pattern. Moreover, since the dataset features annotations of peculiar and unfrequent patterns (such as distractions, recording errors), we can measure the correlation of the captured novelty w.r.t. those. In terms of AUROC, our model scores 0.926, highlighting that novelty can arise from unexpected behaviors of the driver, such as distractions or other shifts in attention. Fig.~\ref{fig:dreyeve} reports the different distribution of novelty scores for ordinary and peculiar events.
\section{Conclusions}
We propose a comprehensive framework for novelty detection. 
We formalize our model to capture the twofold nature of novelties, which concerns the incapability to remember unseen data and the surprisal aroused by the observation of their latent representations. From a technical perspective, both terms are modeled by a deep generative autoencoder, paired with an additional autoregressive density estimator learning the distribution of latent vectors by maximum likelihood principles. To this aim, we introduce two different masked layers suitable for image and video data. We show that the introduction of such an auxiliary module, operating in latent space, leads to the minimization of the encoder's differential entropy, which proves to be a suitable regularizer for the task at hand. Experimental results show state-of-the-art performances in one-class and anomaly detection settings, fostering the flexibility of our framework for different tasks without making any data-related assumption.
\\\\
\textbf{Acknowledgements.} We gratefully acknowledge Facebook Artificial Intelligence Research and Panasonic Silicon Valley Lab for the donation of GPUs used for this research.
{\small
\bibliographystyle{ieee}
\bibliography{references}
}
\clearpage
\twocolumn[
\Large
\begin{center}
\textbf{Supplementary material}
\end{center}
]
\section{On the implementation details}
Architectures and hyperparameters employed for each experiment are reported in Tab.~\ref{tab:model_details}, in terms of the type of blocks, autoregressive layers, mini-batch size, learning rate and weight of the log-likelihood objective. 
All intermediate layers are Leaky ReLU activated. The objective function is optimized using Adam~\cite{kingma2014adam}. All hyperparameters are tuned on a held-out validation set, by minimizing the raw objective (Eq.~\ref{eq:loss} with $\lambda=1$).
\begin{table}[b]
\begin{center}
\resizebox{\columnwidth}{!}{%
\begin{tabular}{ccccccccc}
\toprule
 & MNIST & CIFAR-10 & UCSD Ped2 & ShanghaiTech & DR(eye)VE \\\midrule
\makecell{Input Shape} & 1,28,28 & 3,32,32 & 1,8,32,32\textsuperscript{*} & 3,16,256,512 & 1,16,160,256
\\\midrule
\makecell{Encoder\\Network} & \makecell{D$^{32}_{2,2}$\\D$^{64}_{2,2}$\\FC$^{64}$\\FC$^{64}$}&
\makecell{2D Conv$_\text{3X3}^{32}$\\R$^{32}$\\D$^{64}_{2,2}$\\  D$^{128}_{2,2}$\\D$^{256}_{2,2}$\\FC$^{256}$\\FC$^{64}$}& \makecell{D$^{8}_{1,2,2}$\\ D$^{12}_{2,1,1}$\\ D$^{18}_{1,2,2}$\\  D$^{27}_{2,1,1}$\\ D$^{40}_{1,2,2}$\\TFC$^{64}$} & \makecell{  D$^{8}_{1,2,2}$\\ D$^{16}_{1,2,2}$\\  D$^{32}_{2,2,2}$\\ D$^{64}_{1,2,2}$\\  D$^{64}_{2,2,2}$\\TFC$^{512}$\\TFC$^{64}$} & \makecell{  D$^{8}_{1,2,2}$\\ D$^{16}_{1,2,2}$\\  D$^{32}_{2,2,2}$\\ D$^{64}_{1,2,2}$\\  D$^{64}_{2,2,2}$\\TFC$^{512}$\\TFC$^{64}$} \\\midrule
\makecell{Decoder\\Network} & \makecell{FC$^{64}$\\FC$^{64}$\\ U$^{32}_{2,2}$\\U$^{16}_{2,2}$\\2D Conv$_\text{1x1}^{1}$} & \makecell{FC$^{256}$\\FC$^{256}$\\U$^{128}_{2,2}$\\U$^{64}_{2,2}$\\ U$^{32}_{2,2}$\\R$^{32}$\\2D Conv$_\text{1x1}^{3}$} & \makecell{TFC$^{64}$\\ U$^{40}_{1,2,2}$\\ U$^{27}_{1,2,2}$\\ U$^{18}_{2,1,1}$\\ U$^{12}_{1,2,2}$\\ U$^{8}_{2,1,1}$\\ 3D Conv$_\text{1x1}^{1}$} & \makecell{TFC$^{64}$\\TFC$^{512}$\\ U$^{64}_{2,2,2}$\\ U$^{32}_{1,2,2}$\\ U$^{16}_{2,2,2}$\\ U$^{8}_{1,2,2}$\\ U$^{8}_{1,2,2}$\\3D Conv$_\text{1x1}^{3}$} &\makecell{TFC$^{64}$\\TFC$^{512}$\\ U$^{64}_{2,2,2}$\\ U$^{32}_{1,2,2}$\\ U$^{16}_{2,2,2}$\\ U$^{8}_{1,2,2}$\\ U$^{8}_{1,2,2}$\\3D Conv$_\text{1x1}^{1}$} \\\midrule
\makecell{Estimator\\Network} &
\makecell{MFC$^{32}$\\MFC$^{32}$\\MFC$^{32}$\\MFC$^{32}$\\MFC$^{100}$}&
\makecell{MFC$^{32}$\\MFC$^{32}$\\MFC$^{32}$\\MFC$^{32}$\\MFC$^{100}$}& 
\makecell{MSC$^4$\\MSC$^4$\\MSC$^4$\\MSC$^4$\\MSC$^{100}$}&
\makecell{MSC$^4$\\MSC$^4$\\MSC$^{100}$}&
\makecell{MSC$^4$\\MSC$^4$\\MSC$^4$\\MSC$^4$\\MSC$^{100}$}\\\midrule
\makecell{Mini Batch} & 256 & 256 & 2760 & 8 & 16\\
\makecell{Learning Rate} & $10^{-4}$ & $10^{-3}$ & $10^{-3}$ & $10^{-3}$ & $10^{-3}$ \\
\makecell{$\lambda$} & $1$ & $0.1$ & $0.1$ & $1$ & $1$ \\\bottomrule
\multicolumn{5}{l}{\textsuperscript{*}\footnotesize{Patches extracted from input clips having shape 1,16,256,384.}}
\end{tabular}}
\end{center}
\caption{Architectural and optimization hyperparameters of each setting. We denote with D$^{C}_S$ (downsampling), U$^{C}_S$ (upsampling) and R$^{C}$ (residual) the parametrizations for the employed building blocks (see Fig.~\ref{fig:blocks} in the main paper). On the one hand, $C$ is the number of output channels, whereas $S$ is the stride of the first convolution in the block. Additionally, FC$^{C}$ and TFC$^{C}$ denote dense layers and temporally-shared full connections respectively (in this case, $C$ is the number of output features). Finally, we refer to MFC$^{C}$ and MSC$^{C}$ for the proposed autoregressive layers, illustrated in Fig.~\ref{fig:custom_layers} in the manuscript. For a comprehensive description of each type of layer, please refer to Sec.~\ref{sec:architectural_components} of the main paper.}
\label{tab:model_details}
\end{table}
\section{On the log-likelihood objective}
In this section, we detail how the log-likelihood term (Eq.~\ref{eq:loss} in the main paper) has been computed and optimized. Importantly, as mentioned in the main paper, we model each CPD through a multinomial. To this aim, we firstly need that the encoder acts as a bounded function. To achieve such desideratum, we simply employ a sigmoidal activation, ensuring that latent representations $\textbf{z} = f({\bf x};{\bf \theta}_f)$ reside in $[0,1]^d$. Therefore, for each $z_j$ with $j=1,2,\dots,d$, we perform a linear quantization of the space $[0,1]$ in $B$ bins (where $B$ is a hyperparameter). This latter step provides for $z_j$ a $B$-dimensional categorical distribution $\phi(z_j)$, highlighting the correct bin to which $z_j$ belongs. For each CPD, such distribution will serve as ground truth for the estimator $h({\bf z};{\bf \theta}_h)$, the latter coherently predicting $d$ distributions $p(z_j|\textbf{z}_{<j})$ across the $B$ bins, employing a softmax activation. This way, as shown in Eq.~\ref{eq:llk_loss}, the $\mathcal{L}_{\text{LLK}}$ loss turns out to be a valid likelihood term, defined as the cross-entropy loss between each one of the estimated CPD and their categorical counterparts:
\begin{equation}
\label{eq:llk_loss}
\mathcal{L}_{\text{LLK}}(\theta_f, \theta_h) = \mathbb{E}_{\textbf{x}\sim P}\bigg[-\sum_{j=1}^d\sum_{k=1}^B \phi(z_j)_k \log(p(z_j|\textbf{z}_{<j})_k)\bigg].
\end{equation}
It is worth noting that multinomials are just one of the plausible models for the CPDs. Indeed, if we replace them with Gaussians, the overall framework would leave standing. However, as we observed in different trials, this choice does not yield considerable improvements but rather numerical instabilities, as described in prior works~\cite{oord2016pixel}.

\section{On the relations to Variational Autoencoders}
\label{sec:sup_samples}
\begin{figure*}[h]
\newcommand{\cc}[1]{\multicolumn{1}{c}{#1}}
    \centering
    {
    \begin{tabular}{m{1cm} m{7cm}m{7cm} m{1cm}}
    \toprule
    \cc{FID} & \cc{VAE Samples} & \cc{Our Samples} & \cc{FID}\\\midrule
         \cc{149.72} & \includegraphics[width=0.40\textwidth]{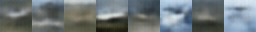} & \includegraphics[width=0.40\textwidth]{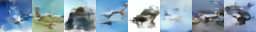} & \cc{72.96}\\
         \cc{172.02} & \includegraphics[width=0.40\textwidth]{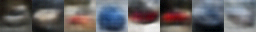} & \includegraphics[width=0.40\textwidth]{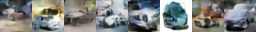} & \cc{72.53}\\
         \cc{181.56} & \includegraphics[width=0.40\textwidth]{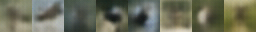} & \includegraphics[width=0.40\textwidth]{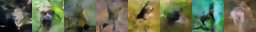} & \cc{76.27}\\
         \cc{188.37} & \includegraphics[width=0.40\textwidth]{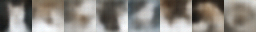} & \includegraphics[width=0.40\textwidth]{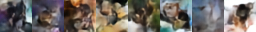} & \cc{67.33}\\
         \cc{202.06} & \includegraphics[width=0.40\textwidth]{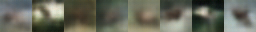} & \includegraphics[width=0.40\textwidth]{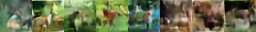} & \cc{68.33}\\
         \cc{207.47} & \includegraphics[width=0.40\textwidth]{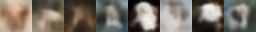} & \includegraphics[width=0.40\textwidth]{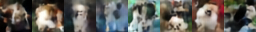} & \cc{73.92}\\
         \cc{186.48} & \includegraphics[width=0.40\textwidth]{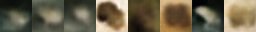} & \includegraphics[width=0.40\textwidth]{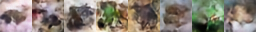} & \cc{62.26}\\
         \cc{220.79} & \includegraphics[width=0.40\textwidth]{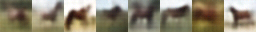} & \includegraphics[width=0.40\textwidth]{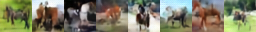} & \cc{64.38}\\
         \cc{164.36} & \includegraphics[width=0.40\textwidth]{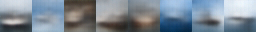} & \includegraphics[width=0.40\textwidth]{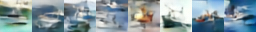} & \cc{52.53}\\
         \cc{204.84} & \includegraphics[width=0.40\textwidth]{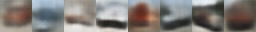} & \includegraphics[width=0.40\textwidth]{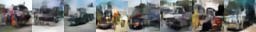} & \cc{67.17} 
    \end{tabular}}
    \caption{For all CIFAR-10 classes (organized in different rows), we report images sampled from VAEs (left) and the proposed autoencoders with autoregressive priors. As can be seen, our samples visually exhibit fine-grained details and sharpness, differently from the heavily blurred ones coming from VAEs. Finally, the over-regularization arising from VAE is confirmed when looking at FID scores (at the extremes of the figure, the lower, the better).}
    \label{fig:samples}
\end{figure*}
Our framework yields some similarities with the Variational Autoencoder (VAE)~\cite{kingma2013auto}. Indeed, they both approximate the integral of Eq.~\ref{eq:lte_integral} in the main paper through the minimization of the reconstruction error under a regularization constraint involving a prior distribution on latent vectors. However, it is worth noting several fundamental distinctions. Firstly, our model does not provide an explicit strategy to sample from the posterior distribution, thus resulting in a deterministic mapping from the input to the hidden representation. Secondly, while VAE specifies an explicit and adamant form for modeling the prior $p(\textbf{z})$, in our formulation its landscape is free from any assumption and directly learnable as a result of the estimator's autoregressive nature. On this point, our proposal leads to two beneficial aspects. First, as the VAE forces the codes' distribution to match the prior, their differential entropy converges to be the same as the prior. This behavior results in approximately stationary entropies across different settings (appreciable in Fig.~\ref{fig:mnist_entropy} in the main paper, where we discuss the intuition behind the entropy minimization within a novelty detection task). Secondly, the employment of a too simplistic prior may lead to over-regularized representations, whereas our proposal is less prone to such risk. Empirical evidence of such behavior can also be appreciated in Fig.~\ref{fig:samples}, where we draw new samples from VAE and our model, both of which has been trained on CIFAR-10. All settings being equal, our hallucinations are visually much more realistic than the ones coming from VAEs, the latter leading to over-smooth shapes and lacking any details, as further confirmed by the substantial differences in Fr\'echet Inception Distance (FID) scores~\cite{heusel2017gans}.
\section{On the dual nature of novelty}
\label{sec:sup_loss_ablation}
\begin{table}[b]
    \begin{center}
        \begin{tabular}{lccc}
        \toprule
        & LLK & REC & NS\\\midrule
        MNIST & 0.926 & 0.949 & \textbf{0.975}\\
        CIFAR-10 & 0.627 & 0.603 & \textbf{0.641}\\
        UCSD Ped2 & 0.933 & 0.909 & \textbf{0.954}\\
        ShanghaiTech & 0.695 & \textbf{0.726} & 0.725\\
        DR(eye)VE & 0.917 & 0.863 & \textbf{0.926}\\\bottomrule
    \end{tabular}
    \end{center}
    \caption{For each setting, AUROC performances under three different novelty scores: i) the log-likelihood term (LLK), ii) the reconstruction term (REC), and iii) the proposed scheme accounting for both (NS).}
    \label{tab:loss_ablation}
\end{table}
In this section, we stress how significant is the presence of both terms for obtaining a highly discriminative novelty score (NS, Eq.~\ref{eq:loss_fusion} in the main paper): namely the reconstruction error (REC), modeling the memory capabilities, and the log-likelihood term (LLK), capturing the surprisal inducted from latent representations. Aiming to reinforce this latter point, just briefly illustrated in Fig.~\ref{fig:loss_fusion} of the manuscript, we report in Tab.~\ref{tab:loss_ablation} performances - expressed in AUROC - delivered by different scoring strategies on each setting mentioned in the main paper. Except for ShanghaiTech, we systematically observe a reward in accounting for both aspects. 
Furthermore, for MNIST and CIFAR-10, we find particularly interesting the gap in performance arising from our reconstruction error w.r.t. the one arising from the denoising autoencoder (DAE) variants (0.942 and 0.590 for the two datasets respectively, as reported in Tab.~\ref{tab:oc_mnist_cifar} of the main paper). In this respect, we gather new evidence supporting that surprisal minimization acts as a novelty-oriented regularizer for the overall architecture, as it improves the discriminative capability of the reconstruction (as already conjectured in Sec.~\ref{sec:one_class_novelty_detection} of the main paper).
\section{On the causal structure of representations}
\begin{figure}[t]
\begin{center}
\includegraphics[width=0.9\columnwidth]{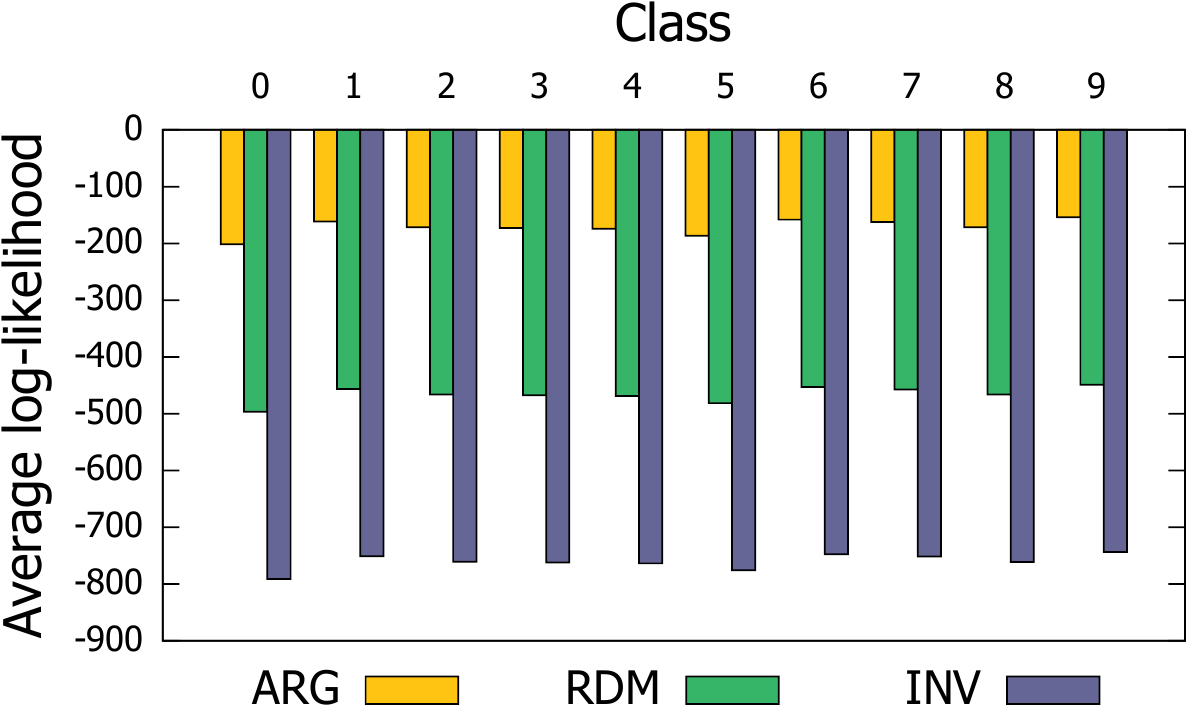}
\end{center}    
\caption{Sample training log-likelihood of a Bayesian Network modeling the distribution of latent codes produced by the encoder of our model trained on MNIST digits. When the BN structure resembles the autoregressive order imposed during training, a much higher likelihood is achieved. This behavior is consistent in all classes and supports the capability of the encoder to produce codes that respect a pre-imposed autoregressive structure.}
\label{fig:bn_likelihood}
\vspace{-.3cm}
\end{figure}
We now investigate the capability of our encoder to produce representations that respect the autoregressive causal structure imposed by the LLK loss (mentioned in Sec.~\ref{sec:model} of the main paper). To this aim, we extract representations out of the ten models trained on MNIST digits and fit their distribution using a structured density estimator. Specifically, we employ Bayesian Networks (BNs) with different autoregressive structures. In this respect, each BN is modeled with Linear Gaussians~\cite{koller2009probabilistic}, s.t. each CPD $p(z_i|Pa(z_i))$ with $i=1,2,\dots,d$ is given by:
\begin{align}
\begin{split}
p(z_i|Pa(z_i)) 
&= \mathcal{N}(z_i \ | \ w^{(i)}_0 + \sum_{\mathclap{z_j \in Pa(z_i)}} w^{(i)}_j z_j,\sigma^2_i),
\end{split}
\end{align}
where each $w^{(i)}_j$, $\sigma^2_i$ are learnable parameters. We indicate with $Pa(z_i)$ the parent variables of $z_i$ in the BN. The previous equation holds for all nodes, except for the root one, which is modeled through a Gaussian distribution. 
Concerning the BN structure, we test:
\begin{itemize}
    \item Autoregressive order: the BN structure follows the autoregressive order imposed during training, namely $Pa(z_i) = \{ z_j \ | \ j = 1,2, \dots, i-1 \}$
    \item Random order: the BN structure follows a random autoregressive order.
    \item Inverse order: the BN structure follows an autoregressive order which is the inverse with respect to the one imposed during training, namely\\$Pa(z_i) = \{ z_j \ | \ j = i+1,i+2, \dots,d \}$ 
\end{itemize}
It is worth noting that, as the three structures exhibit the same number of edges and independent parameters, the difference in their fitting capabilities is only due to the causal order imposed over variables.\\
Fig.~\ref{fig:bn_likelihood} reports the sample training log-likelihood of all BN models. Remarkably, the autoregressive order delivers a better fit, supporting the capability of the encoder network to extract features with learned autoregressive properties. Moreover, to show that this result is not due to overfitting or other lurking behaviors, we report in Tab.~\ref{tab:bn_log_likelihood} log-likelihoods for training, validation and test set.
\begin{figure}[b]
\newcommand{\cc}[1]{\multicolumn{1}{c}{#1}}
\centering
\begin{tabular}{m{1.5cm}m{5cm}}
\toprule
\cc{Loss weight} & \cc{Reconstructions}\\\midrule
\cc{$\lambda=0.01$}&\includegraphics[width=0.6\columnwidth]{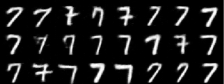}\\
\cc{$\lambda=1$}&\includegraphics[width=0.6\columnwidth]{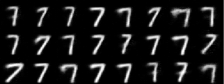}\\
\cc{$\lambda=100$}&\includegraphics[width=0.6\columnwidth]{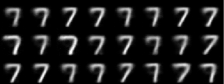}\\
\bottomrule
\end{tabular}\\[1.5ex]
\caption{MNIST reconstructions delivered by different values of $\lambda$, the latter controlling the impact of the differential entropy minimization.}
\label{fig:mnist_entropy_reconstructions}
\end{figure}
\begin{table*}[t]
    \centering
    \resizebox{\textwidth}{!}{%
    \begin{tabular}{|c|c|DDDDDDDDDD|}
    \hline
    \multicolumn{2}{|c|}{}&\multicolumn{10}{c|}{Classes}\\\cline{3-12}
    \multicolumn{2}{|c|}{}&0&1&2&3&4&5&6&7&8&9\\\hline
    \multirow{3}{*}{ARG}&Train&-201.60&-161.60&-171.43&-172.73&-174.17&-186.48&-158.22&-162.37&-171.65&-154.11\\
    &Val&-200.96&-160.38&-170.10&-172.29&-173.85&-185.25&-157.22&-162.20&-171.42&-154.02\\
    &Test&-200.89&-159.73&-169.64&-170.75&-172.40&-184.27&-157.74&-161.65&-170.10&-152.70\\\hline
    \multirow{3}{*}{RDM}&Train&-496.33&-456.34&-466.16&-467.47&-468.90&-481.21&-452.95&-457.10&-466.39&-448.84\\
    &Val&-495.69&-455.11&-464.83&-467.02&-468.58&-479.98&-451.95&-456.93&-466.15&-448.75\\
    &Test&-495.62&-454.47&-464.37&-465.48&-467.13&-479.00&-452.48&-456.38&-464.83&-447.43\\\hline
    \multirow{3}{*}{INV}&Train&-791.06&-751.07&-760.89&-762.20&-763.63&-775.94&-747.68&-751.83&-761.12&-743.57\\
    &Val&-790.42&-749.84&-759.56&-761.75&-763.31&-774.71&-746.68&-751.66&-760.88&-743.48\\
    &Test&-790.35&-749.20&-759.11&-760.22&-761.86&-773.73&-747.21&-751.12&-759.56&-742.16\\\hline
    \end{tabular}}
    \\[1.5ex]
    \caption{Sample log-likelihood obtained by different BN structures when fitting MNIST representations. Each BN is trained on latent codes computed from the training set of a single class, following either the autoregression order (ARG), a random order (RDM) or the order inverse to autoregression (INV). We report the log-likelihood also on the validation and test set. For train-val-test split, see Sec~\ref{sec:one_class_novelty_detection} of the paper. Only \quotes{normal} test samples are used in this evaluation.}
    \label{tab:bn_log_likelihood}
\end{table*}
\section{On the entropy minimization}
To provide an additional grasp about the role of the representation's entropy minimization, we focus on a single MNIST digit (class 7) and report in Fig.~\ref{fig:mnist_entropy_reconstructions} some randomly sampled reconstructions from the training set. Such reconstructions are learned under three different regularization regimes, represented by different weights on the log-likelihood objective ($\lambda$, Eq.~\ref{eq:loss} in the main paper). As shown in Fig.~\ref{fig:mnist_entropy_reconstructions}, higher degrees of regularization (i.e., stricter constraints on entropy) deliver near mode-collapsed reconstructions, losing sharp variations in favor of capturing fewer prototypes for the input distribution.
%
\section{On the complexity of autoregressive layers}
In this section, we briefly discuss the complexity of Masked Fully Connected (MFC) and Masked Stacked Convolution (MSC) layers (Fig.~\ref{fig:custom_layers} of the main paper)\footnote{We refer to the type \singlequotes{B} of both layers, since it is an upper bound to the type \singlequotes{A}}: adhering to the notation introduced in Sec.~\ref{sec:model} from the main paper, MFC exhibits $\frac{d^2+d}{2} \cdot ci \cdot co + d \cdot co$ trainable parameters and a computational complexity $\mathcal{O}(d^2 \cdot ci \cdot co)$. MSC, instead, features $\frac{3d^2+d}{2} ci \cdot co + d \cdot c_o$ free parameters and a time complexity $\mathcal{O}(d^2 \cdot ci \cdot co \cdot t)$. 
%
\section{On the localizations and novelty scores in video anomaly detection}
We show in Fig.~\ref{fig:video_anomaly_detection_suppl} other qualitative evidence of the behavior of our model in video anomaly detection settings, namely UCSD Ped2 and ShanghaiTech.
\bgroup
\begin{figure*}
\centering
\setkeys{Gin}{width=1\textwidth,trim=1.0cm 0cm 1.0cm 0cm,clip}
\resizebox{\textwidth}{!}{
\begin{tabular}{cc}
\includegraphics{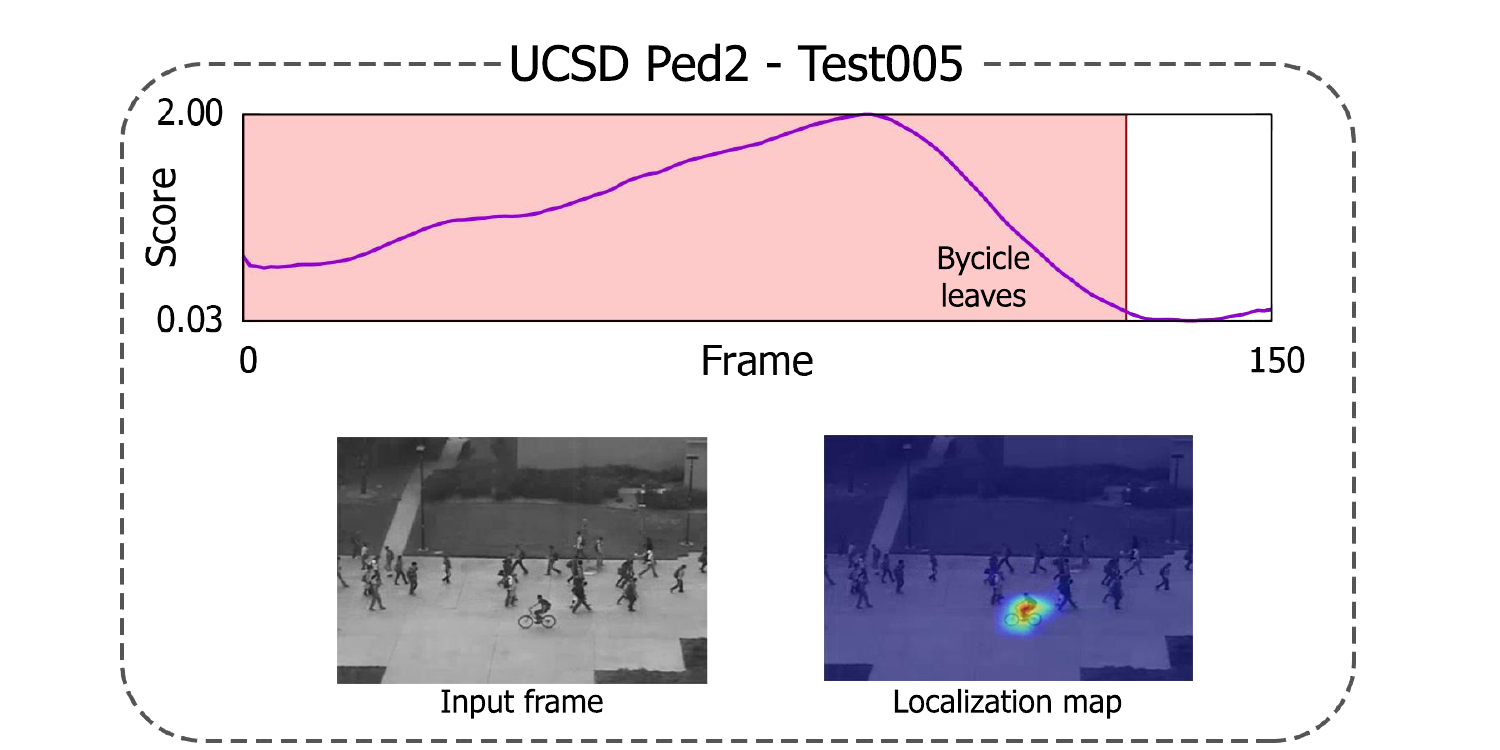}&
\includegraphics{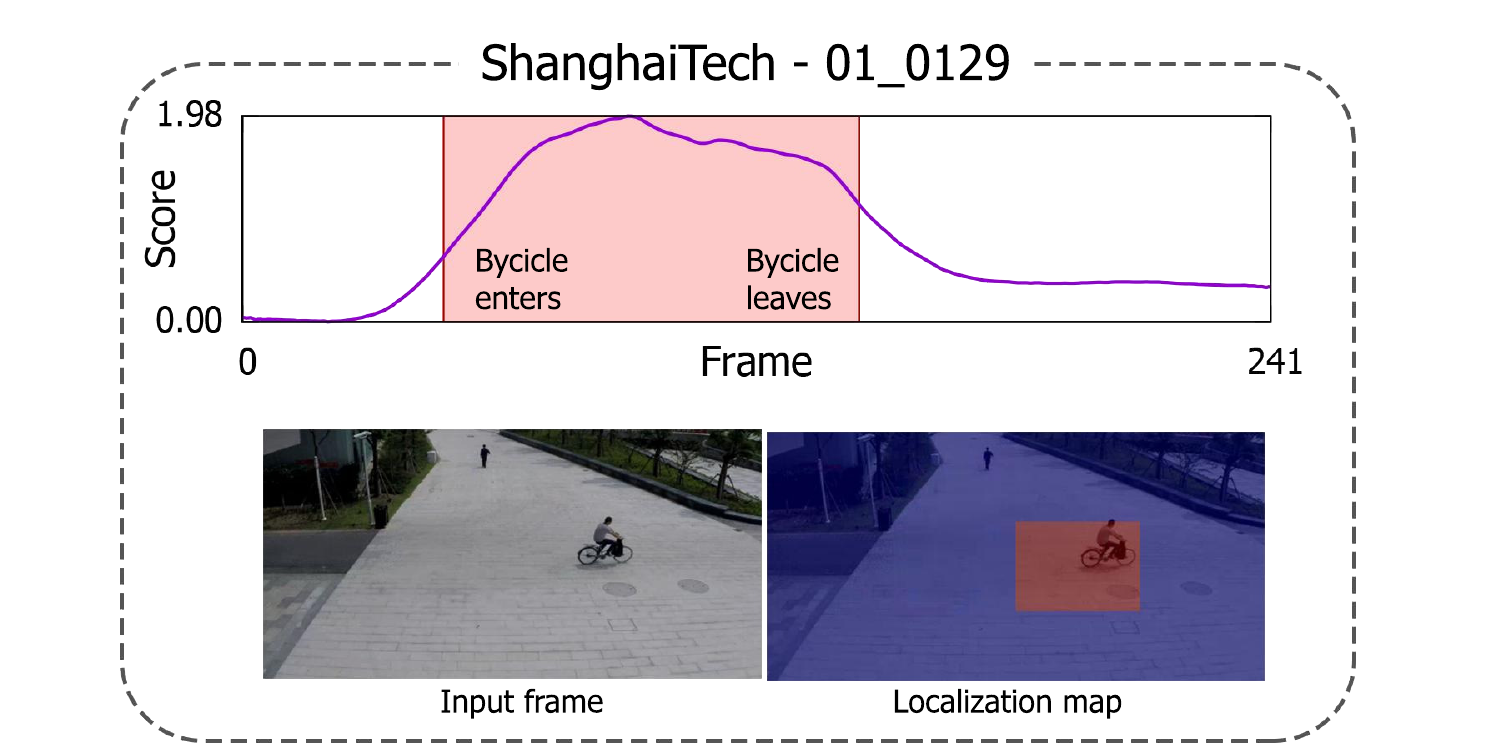}\\
\includegraphics{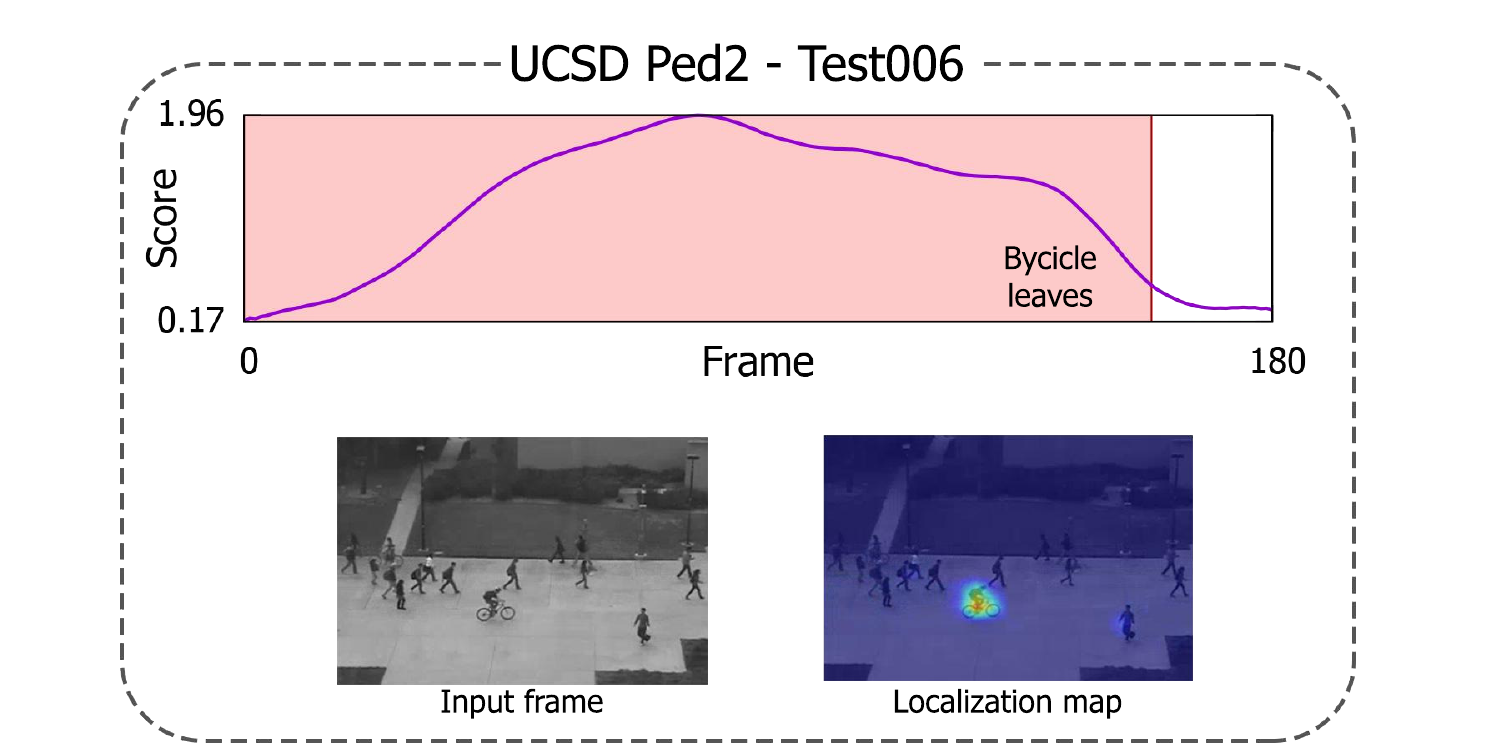}&
\includegraphics{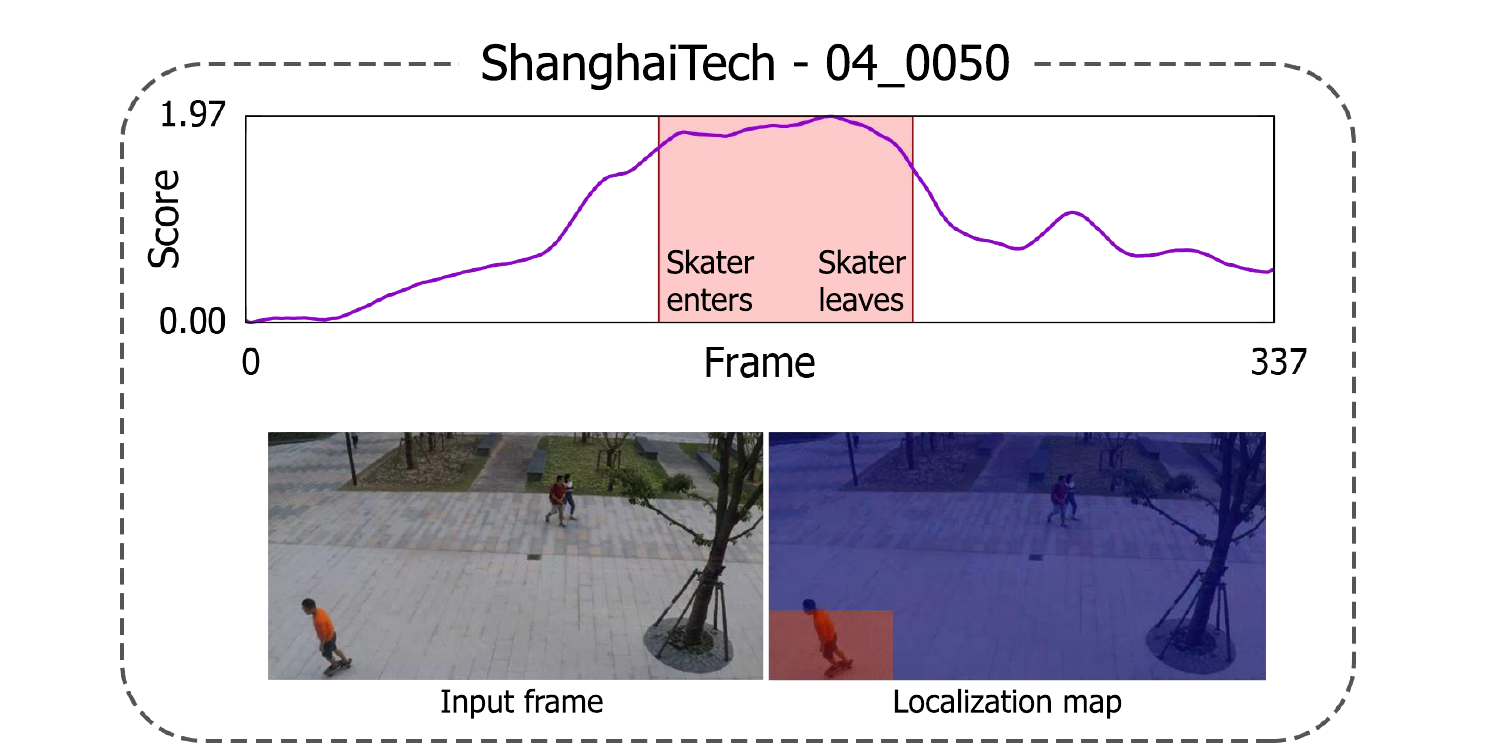}\\
\includegraphics{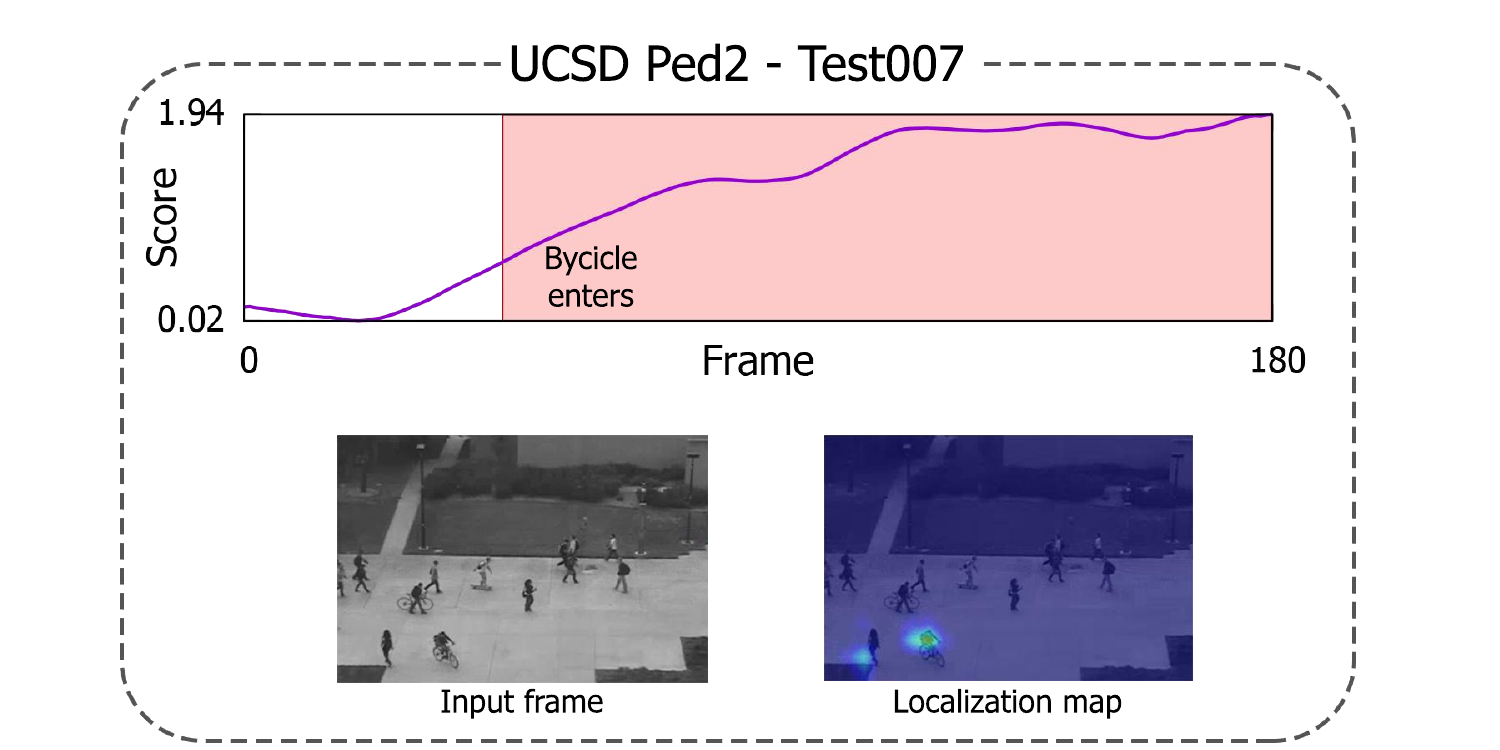}&
\includegraphics{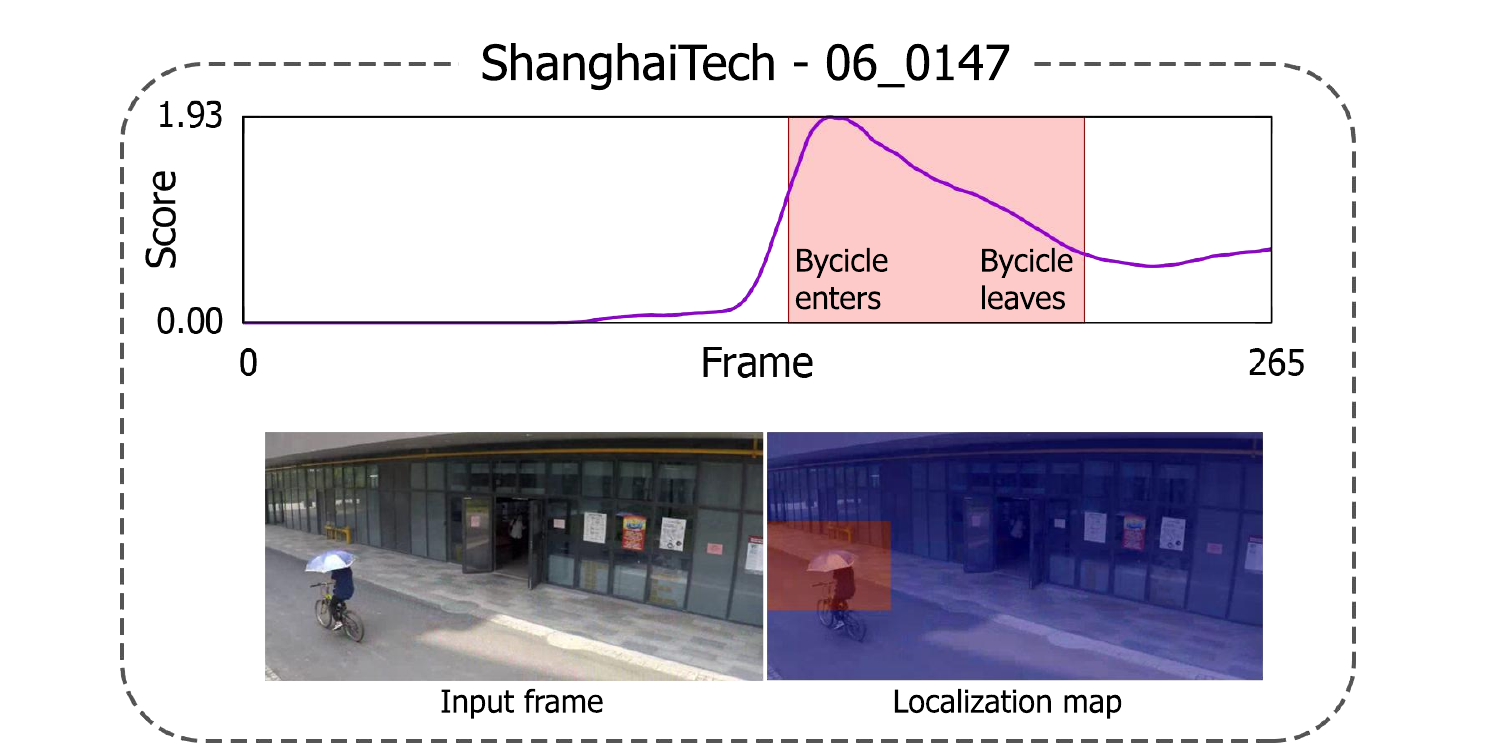}\\
\includegraphics{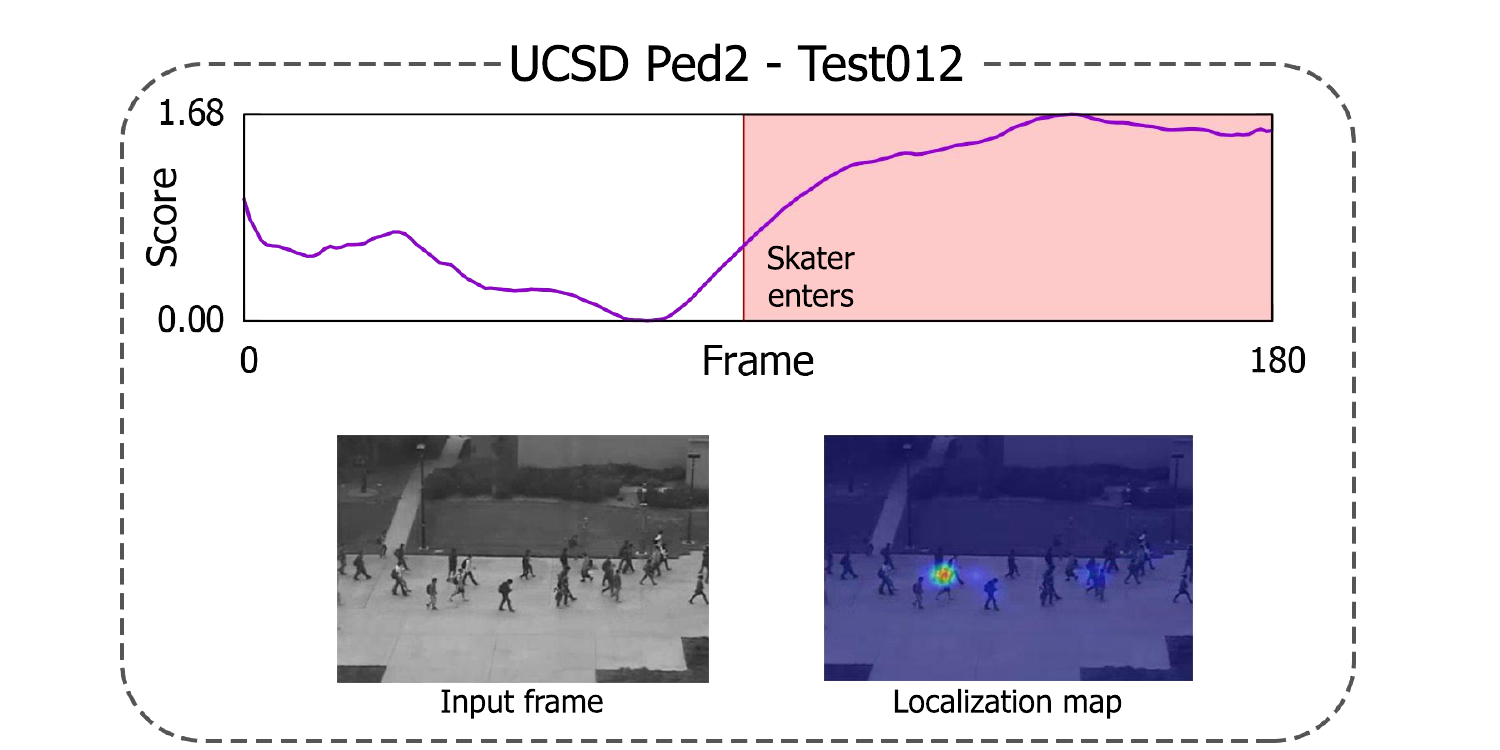}&
\includegraphics{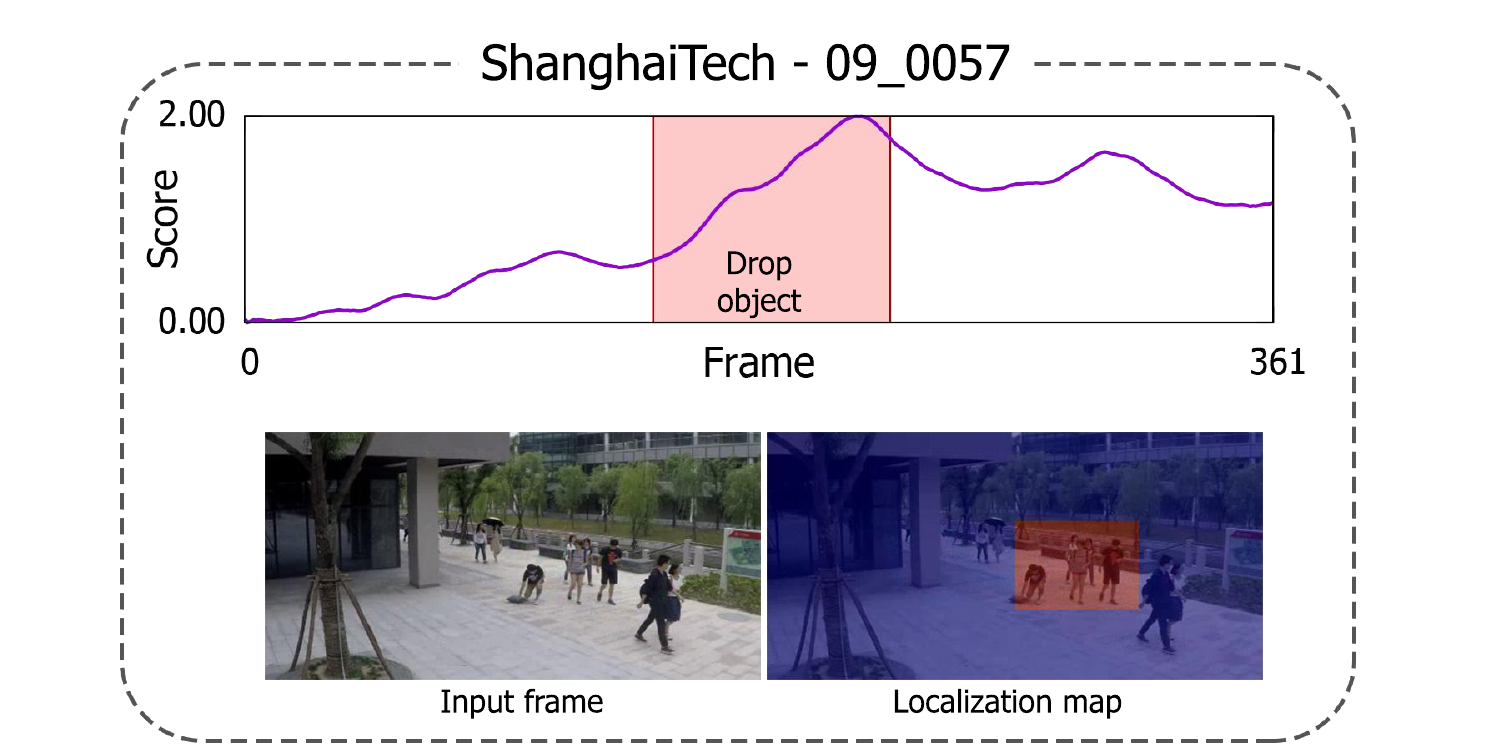}\\
\end{tabular}}\\[1.5ex]
\caption{Novelty scores and localizations maps for several test clips from UCSD Ped2 (left) and ShanghaiTech (right).}
\label{fig:video_anomaly_detection_suppl}
\vspace{-.3cm}
\end{figure*}
\egroup
\end{document}